\documentclass[a4paper, 11pt]{article}

\usepackage{fullpage}
\usepackage{amsmath,amssymb,amsfonts,amsthm}
\usepackage{url}
\usepackage{multirow}
\usepackage{graphicx}
\usepackage{subcaption}
\usepackage{totpages}
\usepackage{comment}
\usepackage[utf8]{inputenc}
\usepackage{xcolor}
\usepackage{booktabs}

\newcommand\crule[2]{\textcolor[HTML]{#1}{\rule{#2}{#2}}}

\newcommand{\grain}{P_\textrm{grain}}
\newcommand{\road}{P_\textrm{road}}
\newcommand{\labels}{P_\textrm{labels}}
\newcommand{\lidar}{LiDAR }
\title{Learning to Detect Fortified Areas}
\author{ 
  Allan Gr{\o}nlund \thanks{Department of Computer Science. Aarhus University. \texttt{jallan@cs.au.dk}.} 
  \qquad 
Jonas Tranberg \thanks{Department of Computer Science. Aarhus University. \texttt{stillwalker1234@live.com}. }
}

\begin{document}
\maketitle

\begin{abstract}
High resolution data models like grid terrain models made from \lidar data are a prerequisite for modern day Geographic Information Systems applications. Besides providing the foundation for the very accurate digital terrain models, \lidar data is also extensively used to classify which parts of the considered surface comprise relevant elements like water, buildings and vegetation. In this paper we consider the problem of classifying which areas of a given surface are fortified by for instance,
  roads, sidewalks, parking spaces, paved driveways and terraces. We consider using \lidar data and orthophotos, combined and alone, to show how well the modern machine learning algorithms Gradient Boosted Trees and Convolutional Neural Networks are able to detect fortified areas on large real world data.
  The \lidar data features, in particular the intensity feature that measures the signal strength of the return, that we consider in this project are heavily dependent on the actual \lidar sensor that made the measurement.
  This is highly problematic, in particular for the generalisation capability of pattern matching algorithms, as this means that data features for test data may be very different from the data the model is trained on. We propose an algorithmic solution to this problem by designing a neural net embedding architecture that transforms data from all the different sensor systems into a new common representation that works as well as if the training data and test data originated from the same sensor. The final algorithm result has an accuracy above 96 percent, and an AUC score above 0.99. 
\end{abstract}





\section{Introduction}
High resolution geographic data models, like digital elevation models made from \lidar measurements, are a prerequisite for modern day
Geographic Information Systems applications like flow modelling and flood risk analysis \cite{digital_terrain, Danner_terrastream, flowacc}.
\lidar data eg. is also used for canopy cover estimation \cite{canopy}, tree type classification \cite{XI20201}, and land cover mapping \cite{lodha_cover_svm}.
In land cover mapping the task is to segment the mapped surface into the parts comprising different phenomena like water, vegetation, and buildings. Manually designing good algorithms for this kind of segmentations is not an easy task. A standard way of computing a segmentation is to cover, or tile, the area in question with a geometric primitive like small rectangles or squares, and then determining for each tile the phenomena occuring in the tile, whether it be road or water, reducing the overall segmentation problem to a (large) list of identical small classification problems. If the tiles are small enough, only one phenomena occur in almost all tiles. This is the approach we consider in this work. Creating high quality classification algorithms for determining the phenomena in a given cell is not an easy problem either,  as it may require several types of measurements, and a relevant representation of a larger area surrounding each tile in question to accurately distinguish between different phenomena. 
For this reason the standard approach is to apply different machine learning classification algorithms from classic Support Vector Machines \cite{lodha_cover_svm}, to  deep learning models from computer vision, such as (convolutional) neural networks \cite{hinton_road, wu_capsule}, to learn the classifiers from the data instead.
The methods vary in exactly what kind of objects they try to distinguish between, the kind of algorithms used for the classification, which  preprocessing is applied, and what kind of data features that are available for the task at hand.
For instance \cite{hinton_road}, uses neural networks and various forms of pretraining and post processing to extract road networks from high resolution images.
In \cite{zhou_svm_intens} the authors propose a method that combines information from a digital elevation model and the intensity of return feature from the \lidar data, and apply Support Vector Machines to learn the final classifier, while \cite{wu_capsule} considers only the digital elevation model, and uses a more recent convolutional neural network architecture to learn the classifier.


In this paper, we consider the problem of locating all the fortified areas like roads, sidewalks, paved driveways, and terraces on any given area of interest.
More formally, our goal is to create an algorithm that can segment a mapped surface into fortified and unfortied areas on country sized data sets.
Such a distinction between fortified and non-fortified areas is interesting in its own right, and has several applications in water flow modelling and planning, as it allows models and analyses to take into account which areas can absorb water and which areas can not.
The end goal of this  work is to create an algorithm that can create fine-grained accurate fortification maps of entire countries,
starting with the country of Denmark that has high quality data freely available.
Such segmentation maps require billions of  measurements to be classified, and the classification needs to be updated whenever new data is collected.
In Denmark, the \lidar data for the entire country is currently updated every five years,
each year updating one fifth, or sometimes when a significant change to the landscape has been made like building a new highway.
While information about road networks can sometimes be found in road network databases, extracting road networks from differend kind of images is still very an active research problem \cite{DBLP:journals/corr/abs-1807-01232}. For Denmark, there are an official road network data base, however this data set only represents  each road segment by poly-lines each with an associated width which is imprecose,
and other fortified areas like pavements, terraces and driveways are not readily available. 

To create a fortification classification algorithm we apply the state of the art machine learning algorithms Gradient Boosted Trees using the LightGBM package \cite{lightgbm} and deep convolutional nets based on the U-Net architecture \cite{unet}. 
The data the algorithms use for the classification task are \lidar measurements, including the return intensity of the LiDAR, and orthophotos, and we consider the data sources both combined and separately.
We note that the digital elevation model of the terrain ($x,y,z$ coordinates) is included in the \lidar measurements. 
The reason for also considering the data sources separately is to be able to apply algorithms in situations when only one of the data sources is available, and while \lidar data may be a richer source of information, up to date orthophotos may often be more readily and freely available.
Furthermore, as we observe and show in Section \ref{sec:data}, the \lidar data feature \emph{Intensity of Return},  from the \lidar mapping of Denmark that we use in this project is heavily dependent on the actual \lidar sensor that made the measurement.
This is highly problematic, in particular for the generalization capabilities of pattern matching algorithms,  as this means that data features for test data may be very different from the data the model is created from.
This is aggravated by the fact that we only have a small local area of Denmark where all the fortified areas has been manually classified, and we need the models we derive from the data from this small area to generalize to the entire country using data created by many different \lidar measurement devices.

\paragraph{Our Apprach and Results}
To handle the data from the different \lidar sensors used for Scanning Denmark, we propose an algorithmic solution based on ideas from neural net density estimation and autoencoders. Our proposed solution transforms data from the different \lidar sensor systems into new common representation that is independent of the actual sensors. We compare the results where the training data  and the test data originate from the same sensor and are not transformed, to the situation where the train data and the test data are transformed data from different sensors, and get the same good results in both cases. 
We show that neural networks for image segmentation, in particular neural networks based on the U-Net \cite{unet} architecture, perform better than the gradient boosted trees.
This is not surprising as we expect that spatial locality and context are important for the problem, properties convolutional neural nets were designed to exploit.
Furthermore, we show that using only one of the available data sources already provides high quality classifications but, expectedly, using both data sources gives the best results.
We test several variations of our  deep learning architecture including recent techniques like \emph{Depth-wise Seperable Convs} \cite{chollet2017xception},  \emph{Dillated convolutions} \cite{holschneider1990real} and various forms of \emph{skip-connections} \cite{he2016deep,zhou2018unet++},  for improving our chosen architecture.
We also test recent neural network architectures designed for working directly with point clouds \cite{pointnet++} such as \lidar
to try and take advantage of the potentially extra information in the \lidar data, that is removed  when the data is transformed into a digital elevation model in rasterized form. However, None of these improve on the results we achieve with the standard U-Net architecture.
Finally, we provide an analysis of what kind of terrain surfaces are most problematic for our algorithms and give an explanation for why this is the case.


\section{Datasets}
\label{sec:data}
In this section we describe the two data sources used in this project.
The first data source is orthophotos containing four feature channels, \emph{red}, \emph{green}, \emph{blue} and \emph{near-infrared}, in $0.12$-meter resolution.
The second data source is Point Cloud data generated with \lidar in $0.15$-meter horizontal resolution and $0.05$-meter vertical resolution.
We have two sets of orthophotos recorded in the spring of 2014 and 2016 respectively and a single set of \lidar Point Cloud data recorded between 2014 to 2015.
Both of our data sources are made freely available by \emph{The Danish Agency for Data Supply and Efficiency}.
We refer to \cite{kortforsyningen} for download instructions.
 
\subsection{LiDAR Features}
The \lidar point cloud data used in this project contain the following relevant features.
\begin{description}
    \item[Position:] The $x$, $y$ and $z$ position of each measurement. The positions are encoded using the reference system EPSG:25832.
    \item[Intensity:] The intensity of the return echo, which is the amount of energy left in a beam after interacting with the surface.
    \item[Scan angle:] The angle between the plane and the point in question, measured in the orthogonal direction to the flight direction. This value only has integer resolution.
    \item[Return number and number of returns:] When light hits semi-transparent materials, such as leafs, multiple returns may be registered.
      \emph{Number of returns} and \emph{Return Number} signify how many returns were registered and where in the list (ordered by return time) the point is.
    \item[LAS Classification:] Classification of the points, using \emph{lasTools} \cite{lastools} heuristic based algorithm. The interesting classes are, \emph{Ground}, \emph{Vegetation}, \emph{Buildings}.
\end{description}

\subsection{Labelled Data}
\label{sec:dat:labelled}
\begin{figure}[ht]
    \centering
    \includegraphics[width=0.45\textwidth]{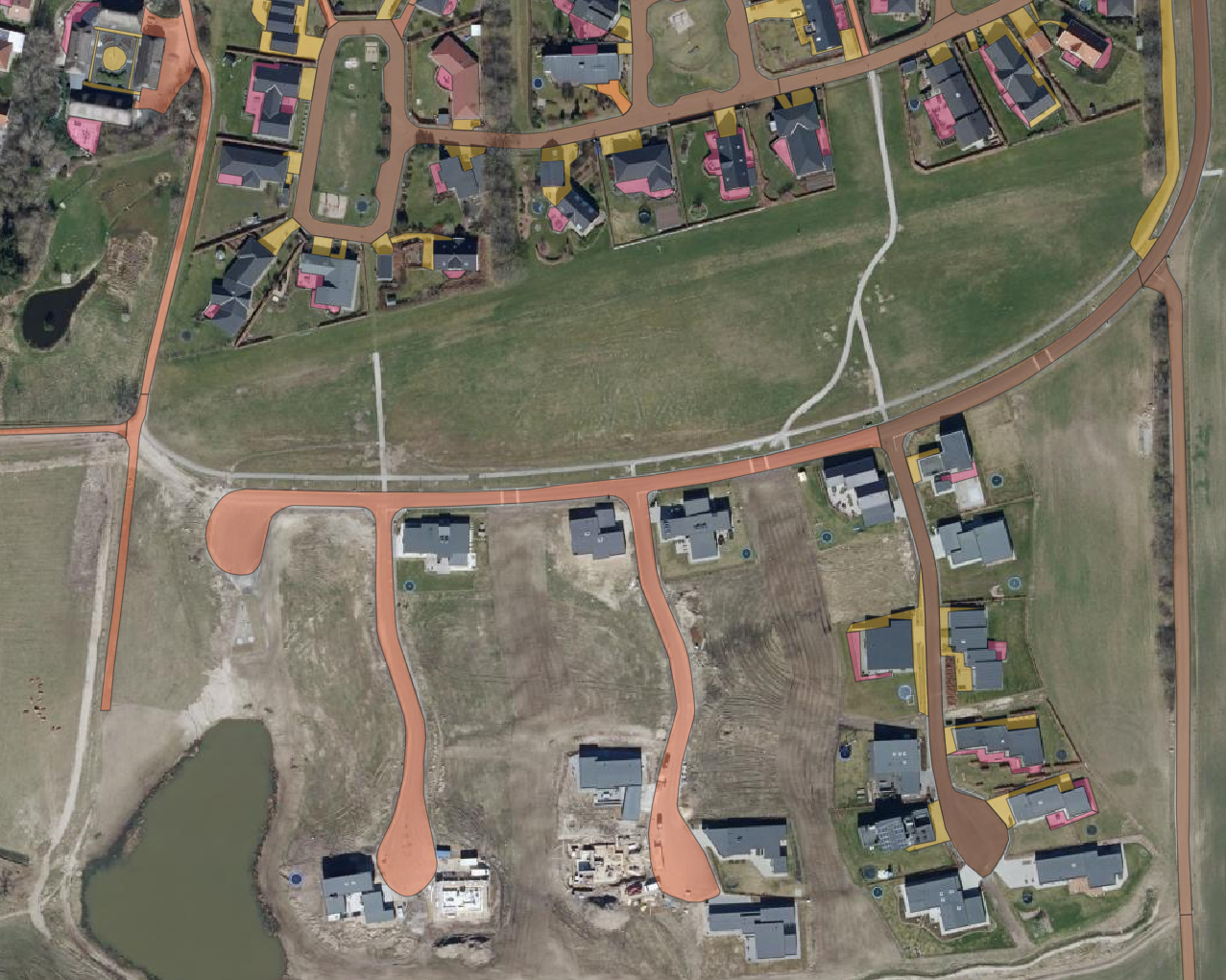}
    \caption{
      Example data from $\labels$. Brown is fortified road, orange is non-fortified road (dirt and gravel road), yellow is fortified sidewalk while pink is fortified terraces.
      This figure exemplifies the issue of time misaligned data sources, as we see that clearly fortified road is marked as non-fortified road.
      Though this example is excluded from training, in general, we expect  errors of this type to be in our training data. 
    }
    \label{fig:skanderborg-labels}
\end{figure}
Though we have an abundance of feature data points, we are lacking in labels. Specifically, we have labels from only a single 
municipality, Skanderborg, in Denmark covering several towns within that area. The labeled data is structured as a set of polygons, denoted $P_\textrm{labels}$,
each polygon marking that the surface area inside the polygon is fortified. This means any area not covered by a polygon in $P_\textrm{labels}$ is a non-fortified area.
The polygons defining the labels were created manually between 2014 and 2016 by the GIS department at Skanderborg Municipality for a similar purpose to ours, and they have graciously allowed us to use them for this project.
The labelled data is the property of the municipality and is not available for download.
The LiDAR data and the orthophotos that our algorithms use for the prediction were  recorded at different times, which means that both of our data sources, as well as the labelled data, is misaligned in some areas.
This is simply due to  natural development of the land areas considered.  An example from the dataset is shown in Figure \ref{fig:skanderborg-labels}, that also illustrates problems with misalignments between the labels and the feature data. 
Besides this set of labels, we have prepared a set of grain fields taken from the entire country of Denmark.  The grain fields are also represented as polygons and we denote this set $\grain$.
The grain fields are also extracted from \cite{kortforsyningen}. Finally, we have created a set of polygons that represent  all public roads in Denmark which we denote by $\road$, also extracted from \cite{kortforsyningen}.
The set of polygons representing the public roads are generated from lines through the center of the roads and lines marking the edges of the roads.
This process is not perfect, and therefore has some significant errors. 

We note that generalizing our algorithms to include the remaining data points and classify them as well is straight forward,
but as our training data only contains buildings as specified by the las tools algorithm, we decided not to include it in our algorithm and analysis to get cleaner results. 

\subsection{Data Specifics}
\label{sec:datasets}
\begin{figure}[ht]
  \begin{subfigure}[t]{0.45\textwidth}
    \centering
    \includegraphics[width=\textwidth, height=5cm]{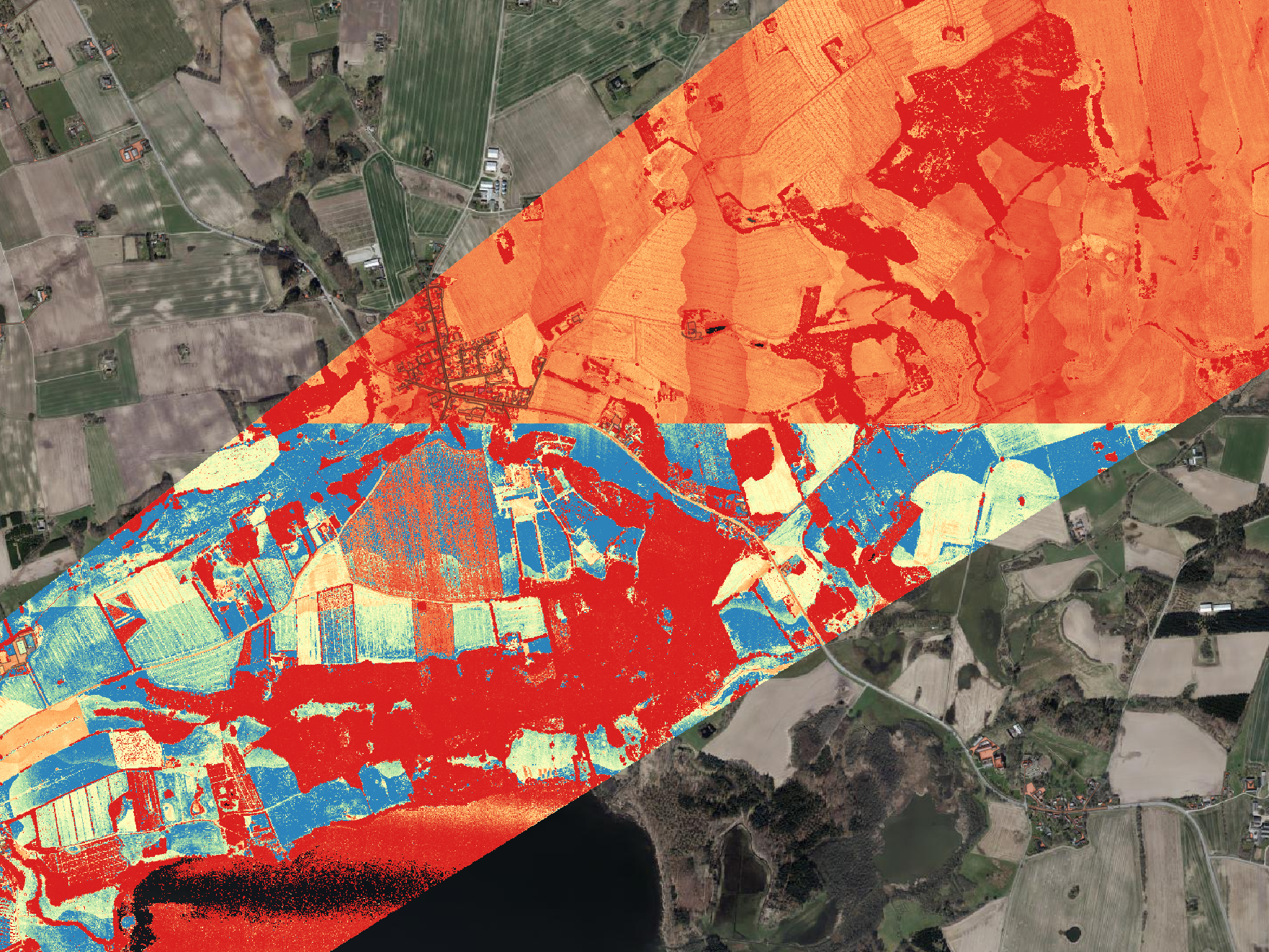}
    \subcaption{Intensity of return \lidar measurements heat-map covering the towns Ry (south of horizontal center) and Skovby (north of horizontal center) in central Jutland. The line that separates the two heat map distributions is also a line defining different sensors used. }
    \label{fig:ry-skovby-intensity}
  \end{subfigure}
  $\quad$  
  \centering
  \begin{subfigure}[t]{0.45\textwidth}
    \centering
    \includegraphics[width=\textwidth, height=5cm]{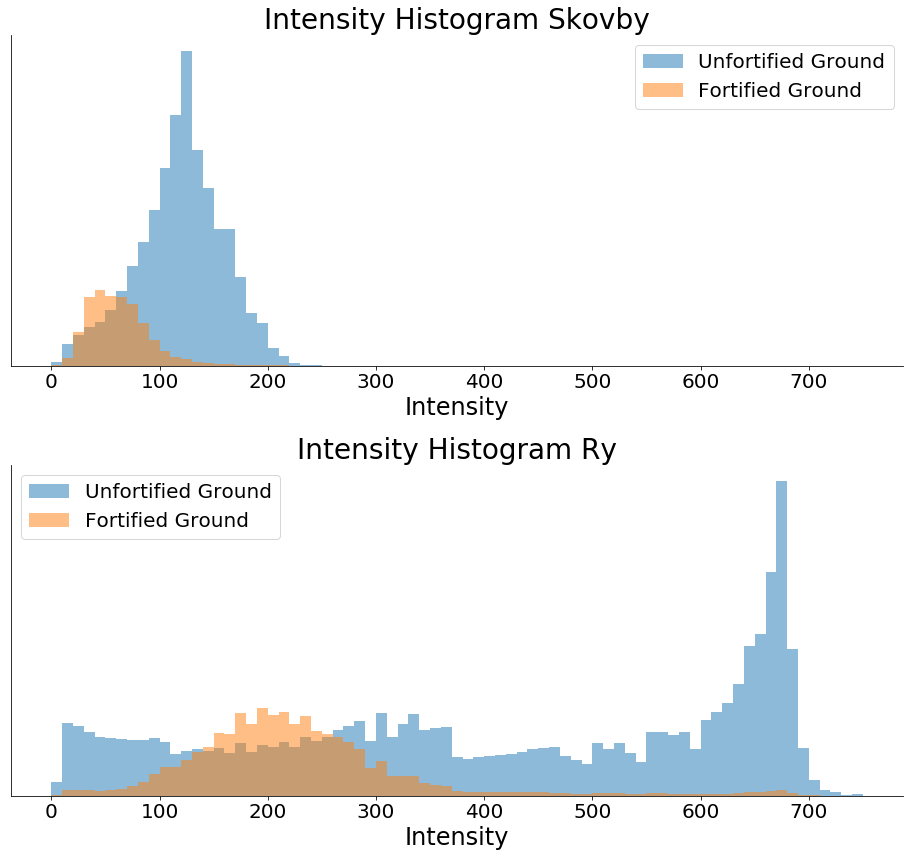}
    \subcaption{Histograms of the intensity of return \lidar measurements from Ry and Skovby. This is the same data used in Figure \ref{fig:ry-skovby-intensity} split into a data set for each of the two sensors used to create the data covering Ry and Skovby respectively. }
    \label{fig:ry-skovby-distribution}    
  \end{subfigure}
  \label{fig:intensity_inputs}
\end{figure}
As we are looking for the fortified areas we are only interested in classyfying \lidar data points actually on the surface.
Therefore, we use the classification made by lasTools to find all the \lidar data points classified as \emph{Ground}.
These Ground data points are the only \lidar data points we try to partition into fortified and unfortified areas.

At first glance, the intensity of return from the \lidar  should be an excellent feature for predicting fortification, since, in contrast to orthophoto features,
intensity is created from controlled light source (the LiDAR instead of the sun) that is independent of variables such as, time of day, time of year and cloud cover etc.
Unfortunately, the \lidar data that we have available, were created with different sensors, resulting in different measurement distributions.
Figure \ref{fig:ry-skovby-intensity} shows the intensity values over a central part of Jutland, clearly demonstrating the different measurement distributions.
Figure \ref{fig:ry-skovby-distribution}, show that this is not merely a shift, but something more complicated is happening.
In Section \ref{sec:methods} we show how to remedy this by learning how to embed intensity data from different sensors into a device independent representation that still contains the relevant information for our task.

\section{Methods}
\label{sec:methods}
This section describes the methods used to create our classification algorithms. We use two supervised learning models. A \emph{Gradient Boosted Tree} model
and a \emph{Convolutional Neural Network}  based on the U-Net architecture.
Finally, we describe how we train a neural network with semi-supervised learning to transform the intensity of return feature from different sensors, into new much more similar representations of the intensity data, 
to account for the differences in the measurement distributions described in Section \ref{sec:datasets}.

\subsection{Gradient Boosted Tree Model}
The Gradient Boosted Tree (\emph{GBT}) is used primarily as a strong baseline, since it is fast to train and very robust to the magnitudes of the input features. The Gradient Boosted Tree models are applied directly to the \lidar (Ground) data points. Formally, a data point is a  vector of fixed dimension $d$ and the goal is to classify whether the data point originates from a fortified or unfortified surface. A GBT predictor is a function from $f: \mathbb{R}^d\rightarrow [0, 1]$, with the interpretation that the output is the estimated probability that the data point is taken from a  fortified surface. Hence predictions near one indicates a data point from fortified surface, and near zero indicates a data point from an unfortified surface.
To get an actual classification we threshold predictions at a half, as is standard.  Since the GBT model has no inherent concept of dependencies between nearby data points,
we append extra features to each data point, created from a local area around the  data point as follows.
For each data point, $p$ in the data set $D$, we define the neighborhood of $p$ as the set of data points $q \in D$ where the Euclidean distance between $p$ and $q$, in the $xy$-plane, is less than $1$ meter.
%
From this neighborhood, the \emph{minimum}, \emph{maximum} and \emph{mean} statistics of the five features, intensity, color and near-infrared are computed and these new features are included in the feature vector for $p$. See Section \ref{sec:experiments} for a more detailed description of these features.
Here it is important to remark that the neighbourhood of a point includes all points from the original data sources,
not just the data points marked as Ground. While we do not wish to classify non-ground points,
presence of non-ground points in a local neighborhood may be relevant information for making accurate predictions.


Furthermore, we compute a \emph{flatness} features from the $x,y, z$ coordinates of the data points in this neighborhood.
This feature is inspired by \emph{pdal}'s \cite{pdal_contributors_2018_2556738} normal filter.
This flatness feature is constructed from the Principal Components of the coordinates $x,y, z$ in the neighbourhood
and is computed as the smallest eigenvalue of the covariance matrix of the data divided by the sum of the three eigenvalues of the covariance matrix of the data. 
A small flatness value for a point $p$, suggest that the neighborhood around $p$ can accurately be represented using only two dimensions that must include the variance in $x$ and $y$ coordinates that exist by construction,
which indicates that the neighborhood around the data point is flat.

Finally, the labels for the data points are created using the label polygon set $P_\textrm{labels}$ the natural way, by labelling a data point fortified if the $x, y$ coordinates of the data point is inside a polygon in $P_\textrm{labels}$ and labelled as unfortified otherwise.

\subsection{The U-Net Model}
The U-Net model \cite{unet} is a well established standard for image segmentation, covered numerous times in the literature and it works as follows.  The input to the U-Net is an image, or 3D tensor, where the first two coordinates index the width and height (the pixels or cells) of the image, and the last coordinate indexes the feature channels of the input tensor.
The output of the U-Net model is a new image/tensor with the same width and height and only one channel, yielding the model prediction for each input pixel.
Let $w, h, d$ be the shape the input tensor, where $w, h$ are width and height respectively and $d$  the number of feature channels,

First, the U-Net model applies $k+1$ layers of \emph{Convolutional Neural Networks, CNN's}, denoted $f^e_i, i=0 \dots k$. Each, $f^e_i$, doubles the number of channels, while a pooling layer downscales the height and width of the output by a factor of 2, effectively doubling the perceptual field of each \emph{CNN} layer. Let $h_i, w_i, c_i$ be the height, width, and number of channels of the tensor output by the $i$'th CNN. We let $e_i, i=0 \dots k$ denote the encoded features after applying the first $i$ CNN transforms, $f^e_i$, and they are defined as follows.
\[
    e_i = 
    \begin{cases}
        f^e_i(I) \in \mathbb{R}^{h_i \times w_i \times c_i} & \text{if } i=0, \\
        f^e_i(p(e_{i-1})) \in \mathbb{R}^{\frac{h_{i-1}}{2} \times \frac{w_{i-1}}{2} \times 2c_{i-1}} & \text{if } i \in 1 \dots k,
    \end{cases}
\]
where $h_0$ and $w_0$ are defined by the size of the input image, and $c_0$ are hyper-parameters, and $p$ is a pooling layer.

After the features $e_0 \dots e_k$ have been created they are upscaled and integrated into the representations at different perceptual scales using a mirror list of \emph{CNN} layers ($f^d_i, i= 0 \dots k$), producing final output $d_0 \in \mathbb{R}^{h_0\times w_0 \times c'}$ for some $c'\geq 1$, as follows:
\[  
    d_i =
    \begin{cases}
        f^d_i(e_i) \in \mathbb{R}^{h_i \times w_i \times c_i}, & \text{if } i = k \\
        f^d_i(e_i \mathbin\Vert u(d_{i+1})) \in \mathbb{R}^{h_i \times w_i \times c_i}, & \text{if } i \in k-1 \dots 0
    \end{cases}
\]
where $u$ is a function that upscale the height and width of a feature tensor using bilinear interpolation, and $\mathbin\Vert$ is concatenation along the $c$ dimension. In all cases $f$ is the following composition of layers:
$$
f = \textrm{Relu} \circ \mathrm{BN} \circ \mathrm{Conv}_1 \circ \textrm{Relu} \circ \mathrm{BN} \circ \mathrm{Conv}_2 \;.
$$
Here $\textrm{Relu}(x)= \max (0, x)$, BN is a \emph{Batch Normalization} layer, and $\mathrm{Conv}_1$ and  $\mathrm{Conv}_2$ are convolutional layers where the width and height of the convolutional filters are always set to three. Finally, our model ends with two, one by one convolution layers separated by a Relu nonlinearity, where the number of channels output by the last layer is one, and that is the output of the entire model. 

We use the  Focal Loss \cite{focal} that generalizes standard cross entropy as the loss function to minimize.
The Focal Loss was designed for image segmentation problems, and allows the learning algorithm to focus on the harder examples.
There may be many easy examples, for instance background pixels in standard image segmentation, or green or yellow fields or similar in our case. Hence, it may be the best strategy for an optimization algorithm to reduce the loss by focusing on the easy examples and improving confidence in predicting these, at the expense of mispredicting the fewer hard examples.

Similarly, as is standard, the training method keeps iterating and improving the model until the validation performance stops improving, and returns the model with maximum validation performance seen in the process.

\subsubsection{Applying U-Net and handling arbitrarily sized regions}
\label{sec:reg_size}
As described above, and different from GBT models, the U-Net model takes as input an image or 3D tensor.
Hence, before we can apply the U-Net model for our problem we need to transform the \lidar data into this format.
To accomplish this, we create rasters from the \lidar data by predefining a grid and for each \lidar data feature mapping the value from all \lidar data points inside a grid cell into a single value. Further details are given in Section \ref{sec:experiments}.
To be able to handle arbitrarily sized and shaped regions with the U-Net model, we follow the construction from \cite{gronlund_corr} that handles this exact problem.
In short, we cover the given input region with fixed size overlapping rectangles, called tiles, matching the expected input size for the U-Net, and use the U-Net model to compute the probability of fortification for each cell in each tile. To compute the output probability of being fortified for each cell, the algorithm computes a weigthed sum of the probabilities computed for the cell for the tiles covering the cell. For this computation the weights depend on the position in the covering tile: The closer to the center of the tile the cell is (more context) the higher the weight and the closer to the boundary the less weight. For a full description of this algorithm we refer to \cite{gronlund_corr}. Figure \ref{fig:heatmap} shows an example where this algorithm combined with our U-Net has been applied to a large area.

\subsection{Transforming Intensity with Embeddings}
To be able to use the intensity feature from different scanners, we need an information preserving transformation of the intensity feature, such that the transformed feature is independent of the LiDAR scanner used.
The idea is as follows.   We assume that the property we are really interested in, the absorption factor of the surface hit, $A$, is transformed through an unknown function $f$ dependent on the \lidar scanner, eg:
$$
I = f_(A, c) \;,
$$
where $I$ is the raw intensity measurement and  $c$ is the context of that measurement (the LiDAR sensor).
In this paper, the context, $c$, refer to the unique \lidar scanner id, but could refer to other variables that influence the measured intensity, such as air pressure, humidity or the angle between the plane and the surface hit.
The goal is to find a new function $f^*$, such that,
$$
A = h(f^*(I, c)) \;,
$$
where $h$ is some invertible function. Notice that we do not need to know $h$, or $A$ which may be hard to find.
 Indeed, we are \emph{only} interested in learning a representation of $A$ that  contains the information about what kind of surfaces the \lidar sensor actually hit, while being  independent of the actual device used for the measurements (context $c$). While we do not have the full labels for any area outside our small training region, we do have the data sets $\road$ and $\grain$ for every region encoding the positions of roads and grain fields, which are fortified and unfortified areas respectively. This means that for each data point we have a label of road, field or unknown and we use these for a semi-supervised approach for learning how to embed the different intensity data distributions into a new common representation.
To realize such a common representation, we employ a basic feed forward neural network inspired by modern neural encoding architectures based on Wasserstein distance like \cite{kolouri2018sliced}
with the difference that, instead of moving the encoded distribution in the direction of some predefined prior,
we want the different input distributions to map to the same encoded distribution, and let the algorithm determine the distribution mapped into as well.

\paragraph{Loss Function}
In the following we describe the loss function we designed to  capture the properties of an encoder we are interested in and train a neural net based encoding algorithm  to minimize that.
The loss function consists of three different parts.
The first part of the loss function is made to make sure that the output distributions from different contexts are as similar as possible as this is vital for any downstream application of the data.
To measure the distance between the encoded intensity distribution under different contexts, we use the \emph{Earth Movers Distance}.
The Earth Mover distance is chosen because it has better gradients for arbitrary distributions than \emph{KL-divergence} \cite{arjovsky2017wasserstein},
and has a practical closed-form solution when the distributions are over the real line, which is the case here.
In general, for distributions over the real line, the closed form solution is (see \cite{ramdas2017wasserstein} for proof):
$$
\textrm{EMD}_1(u, v) = \int_{0}^{1} | U^{-1}(t) - V^{-1}(t) | dt
$$
for distributions $u$ and $v$, with respective cumulative distributions functions $U$ and $V$ (the inverse of these are known as the quantile functions).

We estimate this quantity, for a batch with data from $|c|$ different distributions (contexts) and $n$ samples for each context, by averaging the $\binom{|c|}{2}$ pairwise Earth Mover Distances (or Wasserstein distance) between the $|c|$ distributions.
In total this gives the following loss function
%
$$
\ell_\textrm{emd-all} = \frac{1}{\binom{|c|}{2}} \sum_{i=1}^{|c|} \frac{1}{n} \sum_{j=i+1 }^{|c|} \sum_{k=1}^n |x_{k}^i - x_{k}^j| \;,
$$ 
where $x_k^i$ is the $k$'th largest  embedded intensity value of the $i$'th context, and $n$ is the number of samples considered from each of the contexts.
This is done for all the data points as if no labels were available.
We then exploit the fact that we know field and road data points measured with the different sensors, and add an Earth Mover loss between the field data from different contexts, and between the road data from contexts. These Earth Mover loss  we denote $\ell_\textrm{emd-road}$ and $\ell_\textrm{emd-field}$ respectively.

If there are no other constraints, minimizing this Earth Mover Distances between the encoded distributions is rather simple. An algorithm could simply map all data to zero (or some other constant).
While such a representation solves the task of making encoded distributions from different contexts similar, it is of course useless as the encoded data has no information.
There are several ways we can deal with this problem. The most standard solution is probably to use a decoding algorithm,
and add a reconstruction loss that forces the reconstructed (by the decoder) raw input data  be close to the raw input. This is in essence forces the encoder function to be (approximately) invertible.
However, this is not enough as the encoder may still just map all values to arbitrarily small values close to zero, for instance by setting ($E(x) = x \cdot c$ for $c\rightarrow 0$),  making the encoder function trivially invertible, and the Earth Mover distance between the encoded distributions converges to zero.
We note that this is exactly what a standard implementation of such a setup with neural networks does, and we do not blame it.
Hence, we require an additional part either in the loss function or in the architecture to prohibit this behaviour. One approach is to force a decoder to only use weights with small absolute  value,
for instance by applying weight decay, or one could try and enforce the encoded distributions to have a certain standard deviation by penalizing deviations from some predetermined value.
A more direct solution, and the one we take, is to bound the derivatives of the encoding function. This is a standard method in invertible neural networks like Generative Flow Models for density estimation \cite{DBLP:journals/corr/DinhKB14, Tabak2010DENSITYEB}.
Consider the simplest use case where the encoder function $E$ (for a fixed context $c$) is a mapping  $E: [-1,1] \rightarrow [-1, 1]$.
Then ensuring that $\frac{\partial E}{\partial x} > \varepsilon$ always for some $\varepsilon >0$, ensures that the encoder transformation is invertible,
and that the encoder cannot contract the input arbitrarily close to some fixed point.
Instead of trying to enforce a hard constraint like this, we add the following  penalty depending on the derivatives of the encoder function to the loss function:
$$
\ell_\textrm{grad} = \frac{1}{m} \sum_{i=1}^m  - \lg \left(\frac{\partial E}{\partial x}(x_i)\right) .
$$
Constraining the derivatives like this naturally generalize to higher dimensions by constraining the determinant of the Jacobian of the transformation.
%
%
The final piece of our loss function, is meant to force the embedding function to differentiate between points from fields and roads (unfortified and fortified areas).
There could be many encoding decoding schemes with small earth mover distance, and among these we prefer embeddings that differentiate between the two classes of interest.
To achieve this  we apply a standard linear classification model  to the output of the embedding. We co-train this classifier with the encoding function to distinguish between measurements from roads and fields
using binary cross entropy loss.
This classification loss requires two new trainable parameters $w, b$ which we include in the model.
Formally, the classification loss is defined as follows:
$$
\ell_\textrm{class} = \frac{1}{|c|n} \sum_{i=1}^{|c|} \sum_{k=1}^n \left[ \textrm{BCE}\left( w \cdot E_k^i + b, C_k^i \right) \cdot M_k^i \right],
$$
where \emph{BCE} is the Binary Cross-Entropy function. Here $M_k^i$ is the indicator function, determining whether the $k$'th data point from the $i$'th context is from  a known field or road or whether the labels is unknown, and for the data points where $M_k^i=1$, $C_k^i$ indicates whether that data point is from a field or from a road.
We limit $w$ to be in $[-1, 1]$ as we are not interested in a reduction in loss achieved by scaling $w$.

Adding up the loss functions we arrive at the following combined loss function that we optimize:
%
\begin{equation}
  \label{eq:embed_loss}
\ell = \alpha \cdot (\ell_\textrm{emd-all} + \ell_\textrm{emd-road} + \cdot  \ell_\textrm{emd-field}) +  \beta \cdot \ell_\textrm{grad}  + \gamma \cdot \ell_\textrm{class} \;,
\end{equation}
where $\alpha$, $\beta$, and $\gamma$ are constants weighing the different losses.

\section{Experiments and Results}
\label{sec:experiments}
In this section we report and discuss the results of our experiments.

\subsection{Transforming intensity data experiments}
\label{sec:exp_embedding}
\begin{figure}[ht]
  \centering
  \begin{subfigure}[t]{0.45\textwidth}
    \centering
    \includegraphics[width=\textwidth]{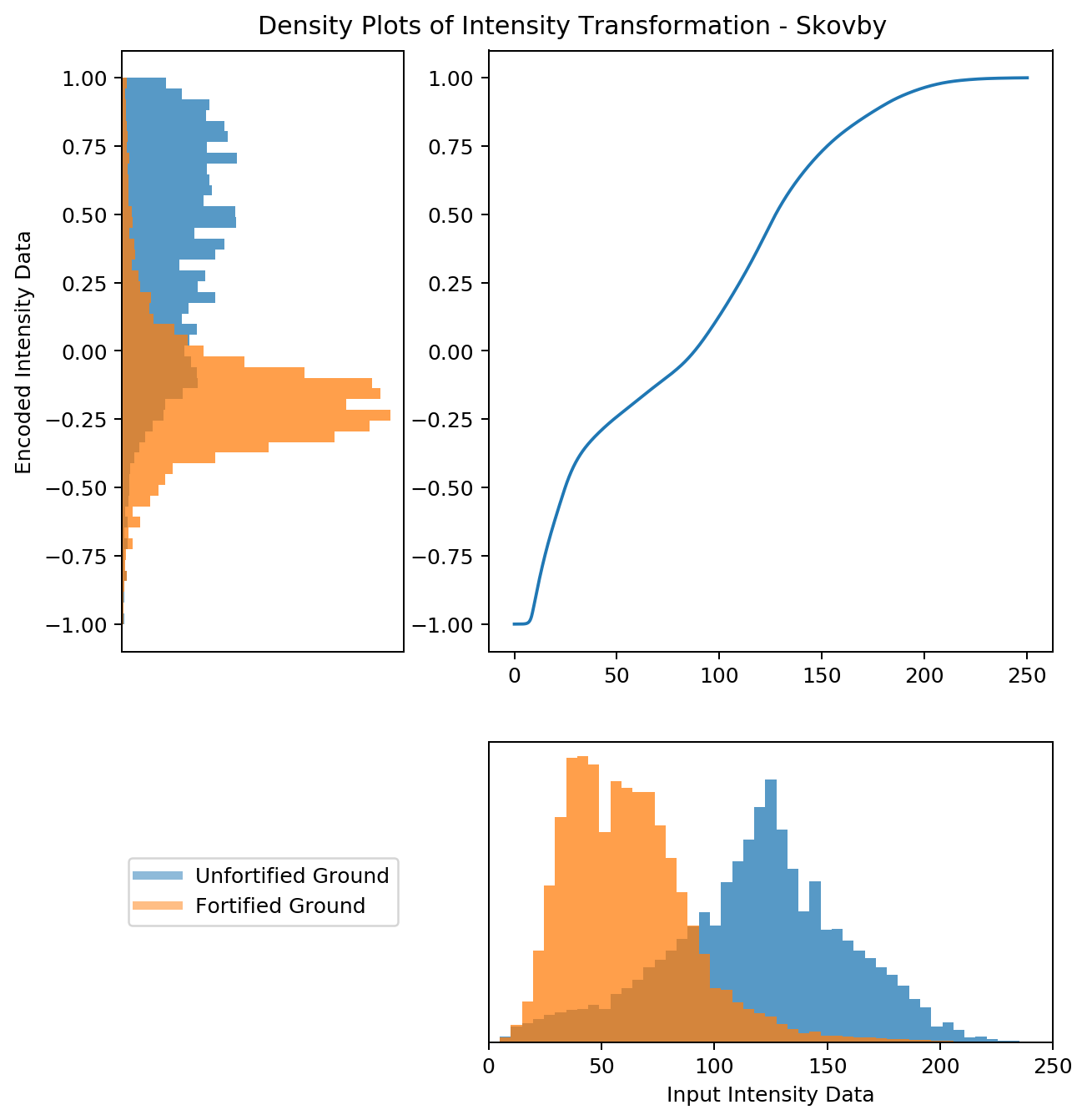}
    \subcaption{Encoding results for Skovby.}
    \label{fig:encoder-dist-skovby}
  \end{subfigure}%
  $\quad$
  \begin{subfigure}[t]{0.45\textwidth}
    \centering
    \includegraphics[width=\textwidth]{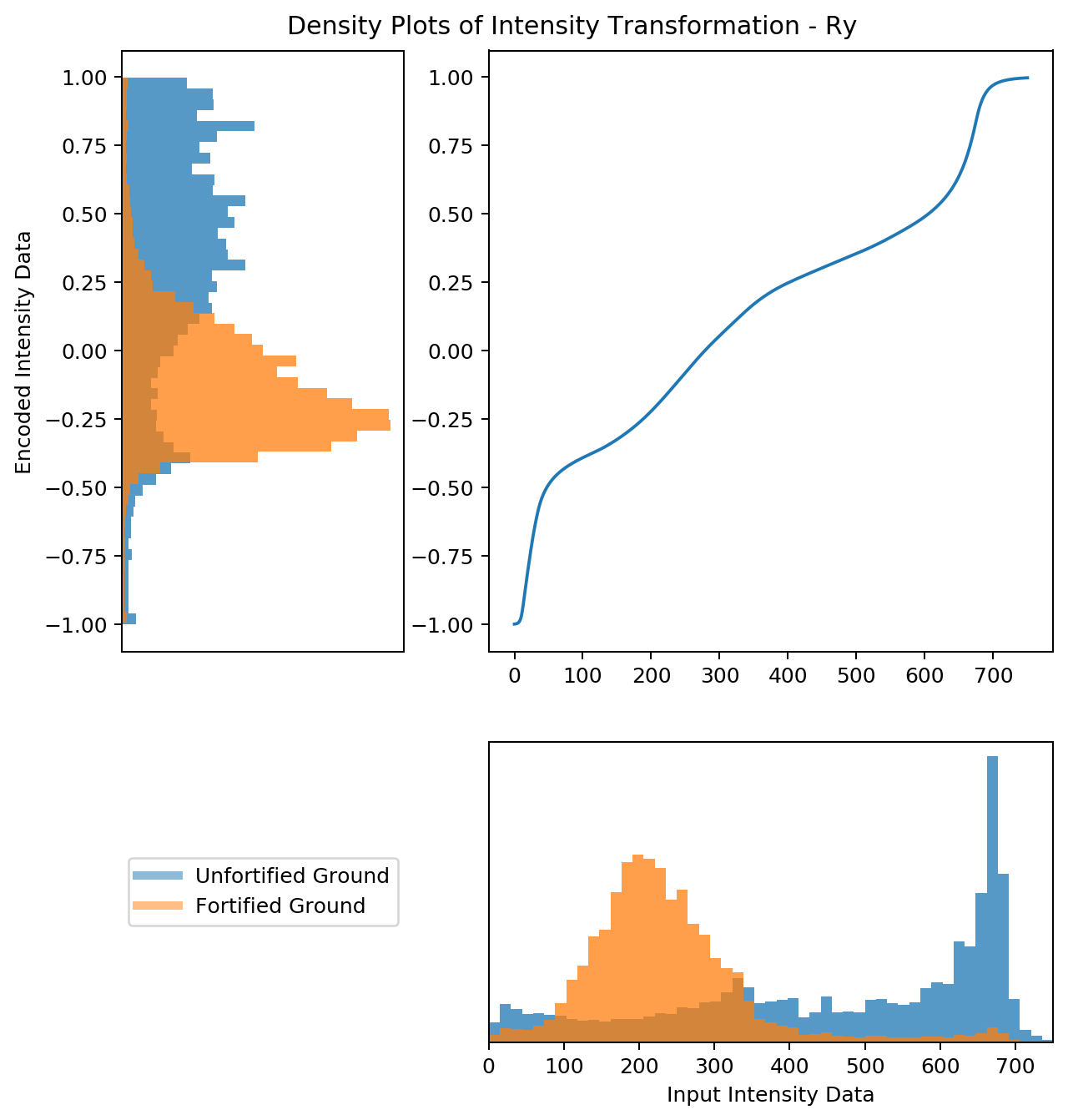}
    \subcaption{Encoding results for Ry.}
    \label{fig:encoder-dist-ry}
  \end{subfigure}
  \caption{ A (normalized to density) histogram of the raw intensity of return \lidar data for Ry and Skovby is shown at the bottom and the histogram for the transformed data is shown above on the left plot,
    while the line plot on the right show learned non-linear transformation the two different LiDAR devices considered in the example.
    To easier understand the visualized data, both the fortified ground and unfortified ground data are shown as density plots, making the two data classes appear to be of equal size which they are not. Notice that the x-axis for the two input distributions are different.
  }     
  \label{fig:intensity_transforms}
\end{figure}
First we discuss our algorithmic approach for transforming the intensity of return values from different \lidar scanners into a new representation independent of the actual scanner.
%
We create a training set for our transformation learning algorithm as follows.
First, a grid of cells, each a square with a side length of one kilometer, covering Denmark is created.
Each cell in the grid is assigned a bucket id, $k$, designating which \lidar sensor is responsible for the measurements in that cell. Cells containing data points from multiple sensors are discarded.
Then, with the set of polygons of grain fields, $\grain$, and the set of polygons of public road segments, $\road$, described in Section \ref{sec:datasets}, we calculate the area of intersection ($I_\textrm{grain}, I_\textrm{road}$ respectively),
of each cell, with respect to both $\grain$ and $\road$ as follows:
$$
I_i(c) = \frac{\sum_{p \in P_i} A(p \cap c)}{A(c)},
$$
where $p_j$ is the $j$'th  polygon in the set of polygons $P_i$ for $i \in \{\textrm{grain}, \textrm{field}$\}, and $A(\cdot)$ is the area function.
Then, we sample cells from the buckets that cover the area we are interested in, such that the number of
 both field and road data points are roughly the same in each LiDAR sensor bucket, while discarding cells that have very few road or field points.
The reason for this is to approximate a situation where the data distribution is the same for each LiDAR sensor bucket,
as otherwise the training procedure might overfit the differences between the different bucket distribution.
Furthermore, during training we oversample road points, so that they always maintain a $20\%$ share of the batch.
From the selected cells we then create the actual training dataset, $D_e = \left\{ (i, c, M, C), \dots \right\}$,
where $i$ is the intensity value,  $c$ is the context id (LiDAR sensor) and $M, C$ are the indicator values for road and grain fields used to define the loss function in Equation \ref{eq:embed_loss}.

Since our training set constitutes the entire population, \emph{out of sample error} do not have much meaning in this context,
and therefore we do not use a validation set, but instead run the learning algorithm until convergence.
For these experiments we use the following weights for the different losses, $\alpha = 1.0$, $\beta = 0.3$, $\gamma = 1.0$.
The behavior of a trained encoder is shown in Figure \ref{fig:encoder-dist-ry}, that visualizes how the trained encoder maps the different input distributions to similar output distributions,
attaining the primary design goal.
In particular, we see that the encoders manage to map the fortified and unfortified ground points, from different contexts, into similar output distributions.
The figure also shows that the encoding functions are  different but still follow the same form of nonlinear pattern and are both (essentially) invertible monotone transformations of the input.

\subsection{Gradient Boosted Trees experiments}
In this section we discuss our experiments with Gradient Boosted Trees for detecting fortified areas using the LightGBM \cite{lightgbm} library.
First,  a standard hyperparameter search is used to  find a reasonable set of hyperparameters. We refer to the LightGBM documentation for an explanation of the many hyperparameters available.
For the experiments, we use a training data set, and a separate validation set, and run the learning algorithm until the performance on the separate validation has not improved over several iterations.
Finally, the model with the highest accuracy on the separate validation set is returned.
We created datasets from three different non-overlapping areas using subsets of the label polygons $\labels$.
Formally, we create a labelled data set by $D = \{ (f_i, l_i), \dots \}$, where $l_i$ is $True$ if the $x, y$ coordinates of the data point falls within any of the polygons in $\labels$ and $f_i$ is vector comprised of the following features:

\begin{itemize}
    \item \emph{Intensity}. Either directly from the LiDAR measurement or from the encoder. This includes both the actual value of the point and the \emph{min}, \emph{max} and \emph{mean} values of the neighborhood as described in Section \ref{sec:methods}. 
    \item \emph{Color \& Nir}. To create these features, we take the 4-band orthophotos and project them onto the LiDAR points using \emph{PDAL}'s \cite{pdal_contributors_2018_2556738} colorization feature.
      This includes both the actual value of the \lidar point and the \emph{min}, \emph{max} and \emph{mean} values of the neighborhood as described in Section \ref{sec:methods}.
    \item \emph{Eigen}. The smallest eigenvalue divided by the sum of the three eigenvalues as computed by a Principal Component analysis on the three dimensional neighborhood of points,
      as described in Section \ref{sec:methods}.
\end{itemize}

Specifically we created datasets $D_\textrm{ry-n}$ , $D_\textrm{ry-s}$ (northern and southern part respectively) and $D_\textrm{skovby}$ sampled in such a way that each data set is label balanced.
As described in the methods section, only  points classified as ground from the LiDAR data are considered, but  all data points are included in the neighborhood statistics.
These datasets are chosen to be geographically separate, to ensure that any correlations between points in close proximity are removed.
Furthermore, the data sets are selected such that the intensity for the data points $D_\textrm{skovby}$ is recorded using a different \lidar scanner than than the data points from $D_{ry-s}$ and $D_\textrm{ry-n}$ which are recorded with the same \lidar scanner.
Finally, to allow comparison of results between the GBT model and the U-Net based model, both the ground-truth polygons defining the labels, and the  predictions made by the GBT model,
are rasterized prior to calculating the results metrics for the GBT model, to match the way both inputs and outputs are rasterized when applying the U-Net model.
More details about the rasterization procedure is given in Section \ref{sec:unet-experiments}.
The target labels are rasterized such that if a point classified as fortified is contained within a raster cell (or a pixel), defined as a bounding-box in the $xy$-plane, that cell is labelled as fortified. 

%
\begin{table}[htb]
  \centering
  \caption{Performance of GBT model trained with different subsets of features. The models are trained on $D_\textrm{ry-s}$ and the results shown for validations sets made from  $D_\textrm{ry-n}$ and $D_\textrm{skovby}$ respectively. }
  \begin{tabular}{cllll}
    \toprule
    Encoder & Validation & AUC & Precision & Recal\\
    \midrule
    & Skovby & 0.8529 & 0.4433 & 0.7892\\
    & Ry-n & 0.9583 & 0.8088 & 0.8442\\
    x & Skovby & 0.9509 & 0.7868 & 0.7995\\
    x & Ry-n & 0.9568 & 0.8034 & 0.8435\\
    \bottomrule
  \end{tabular}
\label{table:gbm-encoder}
\end{table}
%
Our first experiment tests whether the performance of the model is affected by the change in LiDAR scanner and whether the encoder can remedy this. For this experiment we created data sets of 32-thousand data points.
The results are shown in Table \ref{table:gbm-encoder}.
We see that without encoding the intensity feature, performance is greatly decreased in $D_\textrm{skovby}$ where the data was recorded with a different \lidar scanner,
compared to the performance on the area $D_\textrm{ry-n}$ where the data was recorded with the same \lidar scanner that was used on the training data $D_\textrm{ry-s}$.
On the other hand, when using the encoder, performance between the two validation sets are essentially the same.
\begin{table}[htb]
    \centering
    \caption{Performance for GBT models trained on $D_\textrm{ry-s}$ and validated on $D_\textrm{ry-n}$ and $D_\textrm{skovby}$.
      F1 is F1 score and MCC is Matthews Correlation Coefficient (also known as  Pearson's phi coefficient).
    }
    \begin{tabular}{lcccrrrrr}
\toprule
& Color \& Nir & Intensity & Eigen  &   F1  &   AUC  &   MCC  &   Precision  &   Recall  \\
\midrule
Ry-n &  &   &   &                  0.378  &            0.464  &                -0.058  &                  0.274  &               0.610 \\
Ry-n &  &   &  x &                  0.450  &            0.590  &                 0.141  &                  0.357  &               0.608 \\
Ry-n &  &  x &   &                  0.769  &            0.929  &                 0.673  &                  0.766  &               0.772 \\
Ry-n &  &  x &  x &                  0.795  &            0.940  &                 0.713  &                  0.813  &               0.778 \\
Ry-n & x &   &   &                  0.771  &            0.933  &                 0.670  &                  0.710  &               0.844 \\
Ry-n & x &   &  x &                  0.781  &            0.940  &                 0.684  &                  0.726  &               0.846 \\
Ry-n & x &  x &   &                  0.811  &            0.955  &                 0.730  &                  0.785  &               0.838 \\
Ry-n & x &  x &  x &                  0.827  &            0.961  &                 0.754  &                  0.815  &               0.839 \\
\midrule
Skovby &  &   &   &                  0.311  &            0.502  &                 0.008  &                  0.211  &               0.588 \\
Skovby &  &   &  x &                  0.235  &            0.480  &                -0.026  &                  0.193  &               0.301 \\
Skovby &  &  x &   &                  0.690  &            0.916  &                 0.603  &                  0.631  &               0.762 \\
Skovby &  &  x &  x &                  0.714  &            0.923  &                 0.635  &                  0.682  &               0.749 \\
Skovby & x &   &   &                  0.748  &            0.928  &                 0.681  &                  0.736  &               0.761 \\
Skovby & x &   &  x &                  0.759  &            0.934  &                 0.695  &                  0.758  &               0.759 \\
Skovby & x &  x &   &                  0.786  &            0.956  &                 0.729  &                  0.788  &               0.783 \\
Skovby & x &  x &  x &                  0.806  &            0.963  &                 0.758  &                  0.832  &               0.783 \\
\bottomrule
    \end{tabular}
    \label{table:gbm-check-inten}
\end{table}
The second experiment tests how the encoded \emph{intensity} affects the performance relative to the other features.
For this experiment we created data sets of size 128-thousand data points and ran each experiment three times. The results are shown in Table \ref{table:gbm-check-inten}. The result shows that, overall, \emph{Color \& Nir} is the most important feature, but the encoded \emph{Intensity} feature improves performance across the board. The experiment also shows that the performance on data from  the separate town Skovby matches the performance on the data taken from the same town as the training data.

\subsection{U-Net experiments}
\label{sec:unet-experiments}
In this section we present our experiments with Neural Networks for image segmentation, in particular the U-Net architecture.
For this part we have created rasterized data sets from the data sources with a $0.2m \times 0.2m$ resolution with the following features:
\begin{itemize}
    \item \emph{LiDAR}. This feature group comprises of \emph{Ground}, \emph{Return Num} and height \emph{z}. Here, \emph{Ground} specifies whether only points classified as ground are present in the pixel. \emph{Return Num} specify the maximum return number for the points in the pixel, signifying places where canopy overhang the ground. Lastly, \emph{z} is the mean height of the points in the pixel. We normalize the \emph{z} values per tile.
    \item \emph{Color \& Nir}. Same as in the GBT experiment, but rasterized prior to input. These features are rasterized using the mean value of the points inside the pixel box.
    \item \emph{Intensity}. Same as above, though only points classified as ground are rasterized. 
\end{itemize}
For the experiments the U-Net model is trained on tiles of size $96 \times 96$ pixels, cropped from rasterized versions of the point cloud datasets: $D_\textrm{ry-n}$, $D_\textrm{ry-s}$ and $D_\textrm{skovby}$.  We implemented our U-Net architecture in \emph{pytorch} and used the \emph{Ranger} algorithm \cite{liu2019variance, zhang2019lookahead} as our training optimizer.
The first \emph{U-Net} model we consider uses three layers of downsampling and upsampling, which results in a model with about $3.6$ million parameters.
The result of training this model on $D_\textrm{ry-s}$ and validated on $D_\textrm{skovby}$, using encoded Intensity, is given in Table \ref{table:cnn-check-inten}. We see that the U-Net model performs better overall than the \emph{GBT} models. 
\begin{table}[ht]
  \caption{Performance for \emph{U-Net} models using different subsets of available features.
    The models are trained on $D_\textrm{Ry-s}$ and the results shown for the validation set from $D_\textrm{Skovby}$. The values are the mean over eight independent runs of each feature combination.}
  \centering
  \begin{tabular}{cccrrrrr}
\toprule
Color-nir & Intensity & Lidar & F1 Score &   AUC &   MCC & Precision & Recall \\
\midrule
&   &   &     0.279  & 0.705  & 0.201  &     0.441  &  0.206 \\
&   &  x &     0.709  & 0.920  & 0.640  &     0.680  &  0.741 \\
&  x &   &     0.803  & 0.964  & 0.758  &     0.805  &  0.801 \\
&  x &  x &     0.886  & 0.986  & 0.859  &     0.860  &  0.913 \\
x &   &   &     0.805  & 0.970  & 0.760  &     0.779  &  0.834 \\
x &   &  x &     0.875  & 0.985  & 0.846  &     0.839  &  0.914 \\
x &  x &   &     0.826  & 0.975  & 0.786  &     0.800  &  0.856 \\
x &  x &  x &     0.886  & 0.987  & 0.859  &     0.851  &  0.922 \\
\bottomrule
  \end{tabular}
  \label{table:cnn-check-inten}
\end{table}

\subsubsection{Training with more data}
A pressing question is whether more training data can improve the performance of the \emph{U-Net} model.
To test this we increase the data size by a factor of approximately 16 by including data from the city Skanderborg to the training data.
For this large data set, we also consider increaing the number of layers of up and down sampling in the U-Net model from three to five drastically increasing the number of parameters by approximately a factor 20. The results are shown in In Table \ref{table:large_test}.
As we can see, more data does improve performance, however compared to the massive increase in data size the performance increase is modest.
Furthermore, the results also shows that using a larger U-Net network on the large data yields another modest increase in performance across all statistics.

\begin{table}
  \caption{Comparing results for different network and training data sizes based on five independent runs of each layer, data combination.
    Layers is the number of layers used in the U-Net model.}
  \centering
  \begin{tabular}{lllllllll}
    \toprule
Layers & Data Size & Statistic & Accuracy & F1 & AUC & MCC & Precision & Recall \\
\midrule      	     	       	 	    	      
\multirow{2}{*}{3} & \multirow{2}{*}{Small} & Mean & 0.9577 & 0.8884 & 0.9876 & 0.8625 & 0.8730 & 0.9044   \\
                                          &  & Max & 0.9607 & 0.8947 & 0.9899 & 0.8705 & 0.8910 & 0.9214   \\
\multirow{2}{*}{3} & \multirow{2}{*}{Big}   & Mean & 0.9634 & 0.9056 & 0.9900 & 0.8840 & 0.8714 & 0.9426   \\
                                          &  & Max & 0.9642 & 0.9073 & 0.9901 & 0.8860 & 0.8755 & 0.9477   \\
\multirow{2}{*}{5} & \multirow{2}{*}{Big}   & Mean & 0.9662 & 0.9124 & 0.9910 & 0.8924 & 0.8821 & 0.9450   \\
                                          &  & Max & 0.9684 & 0.9173 & 0.9914 & 0.8983 & 0.8936 & 0.9552   \\
\bottomrule
  \end{tabular}
  \label{table:large_test}
\end{table}


\subsubsection{Analysis of Errors}
\label{sec:unet_errors}
\begin{figure}
    \centering
    \begin{minipage}[t]{0.3\textwidth}
        \centering
        \includegraphics[width=\textwidth]{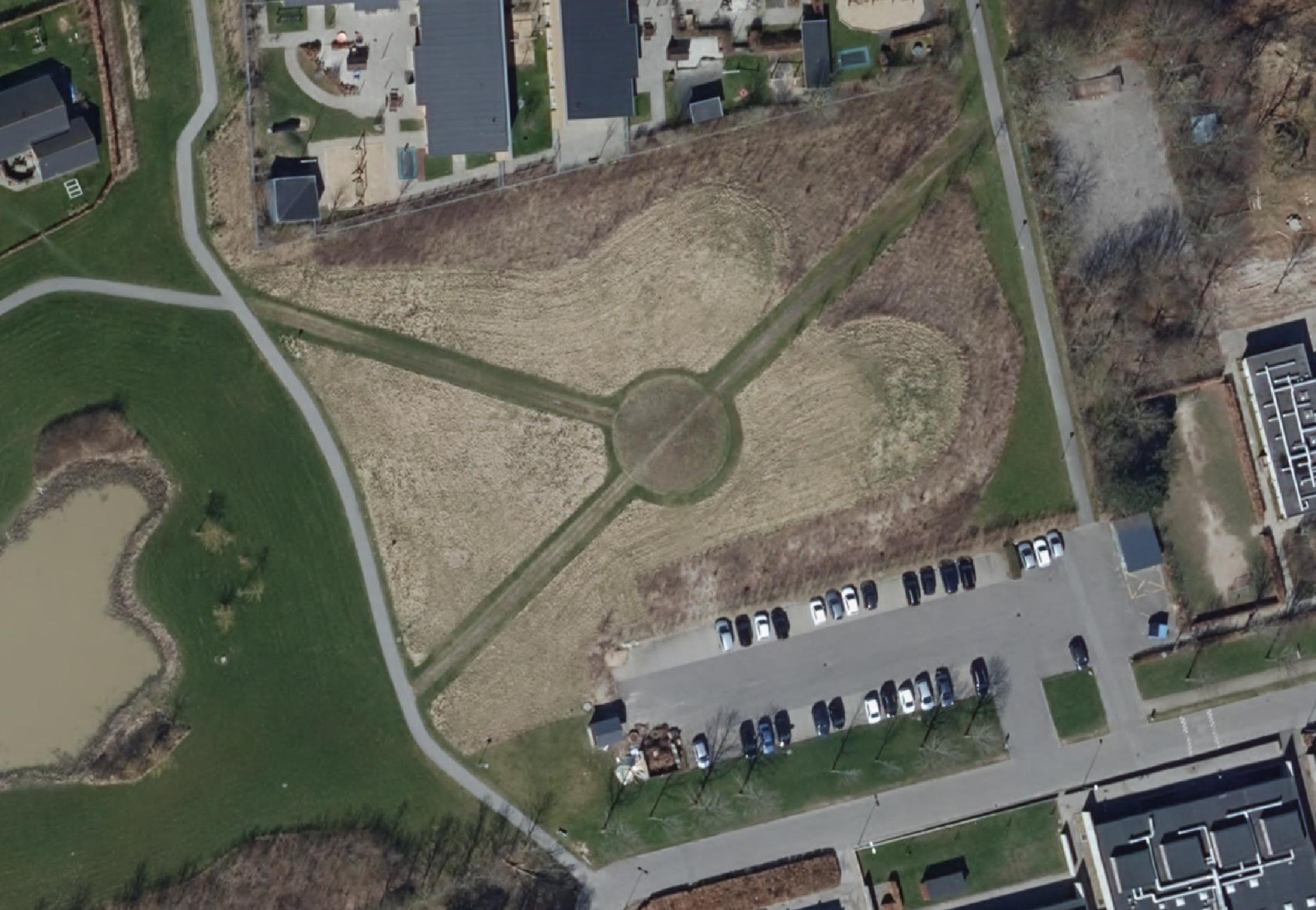}
        \includegraphics[width=\textwidth]{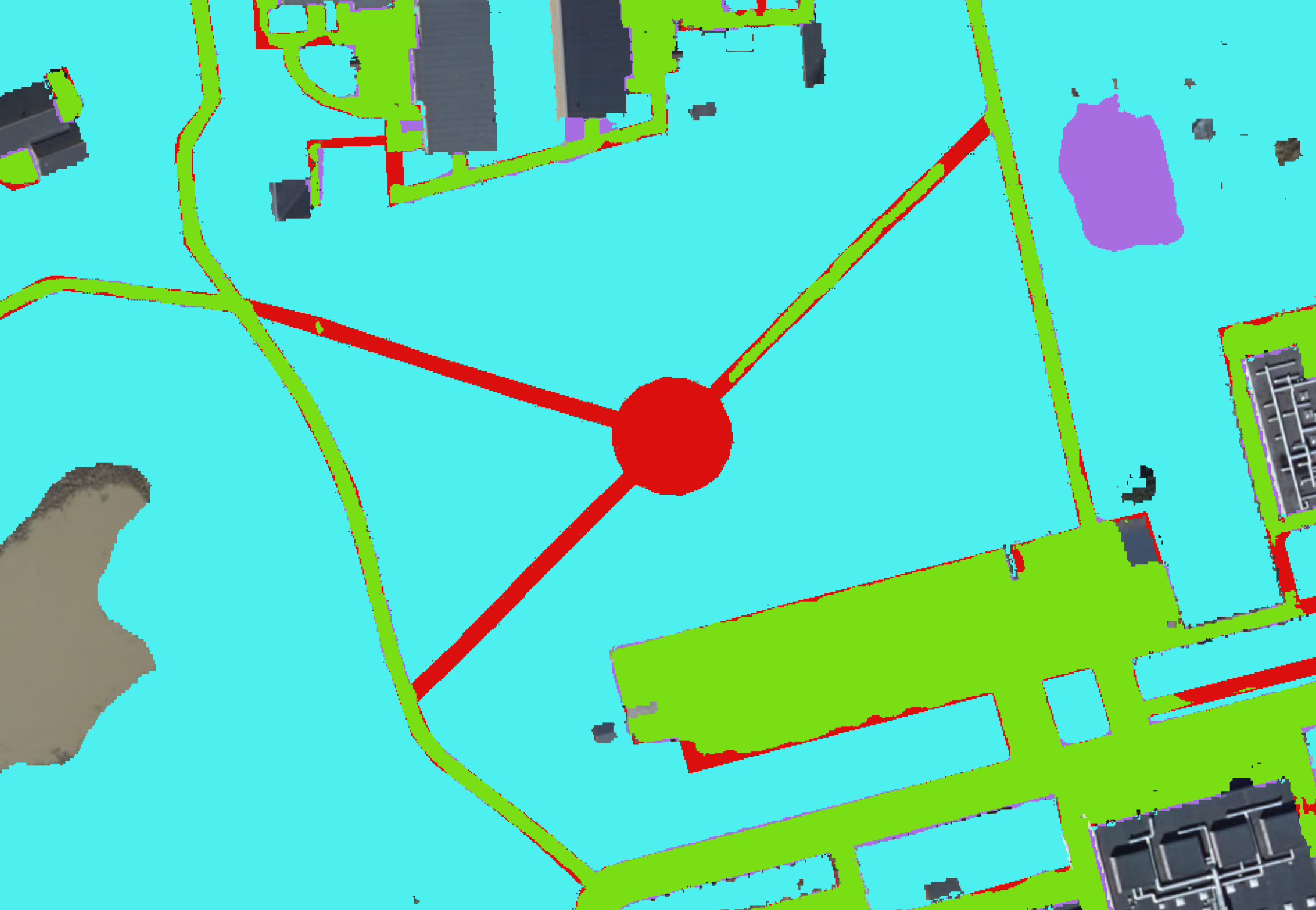}
        \subcaption{Example of bad labels. The grass covered path in the center might have been fortified at some point, but there is nothing in the input features to suggest so.}
    \end{minipage}
    \quad
    \begin{minipage}[t]{0.3\textwidth}
        \centering
        \includegraphics[width=\textwidth]{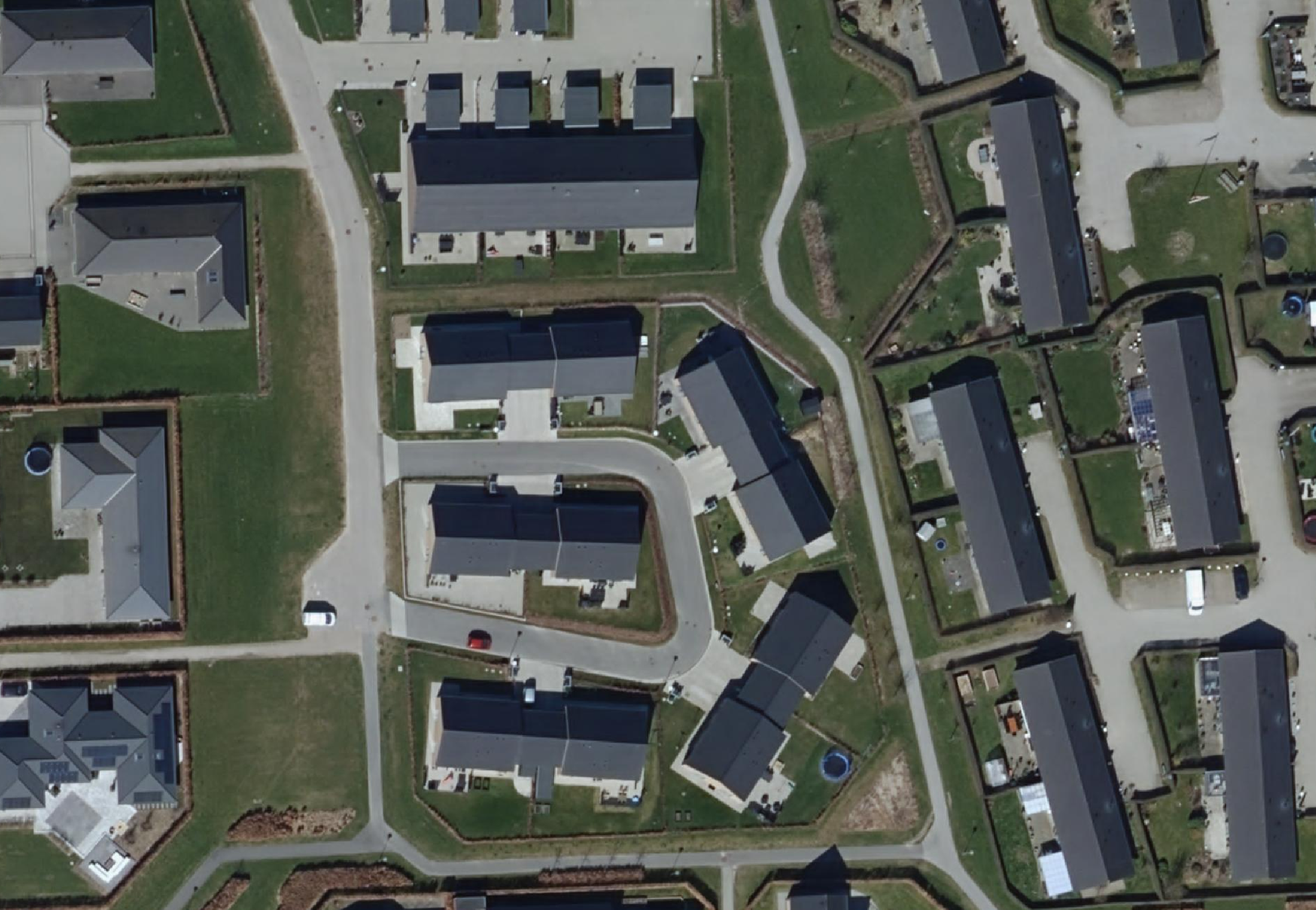}
        \includegraphics[width=\textwidth]{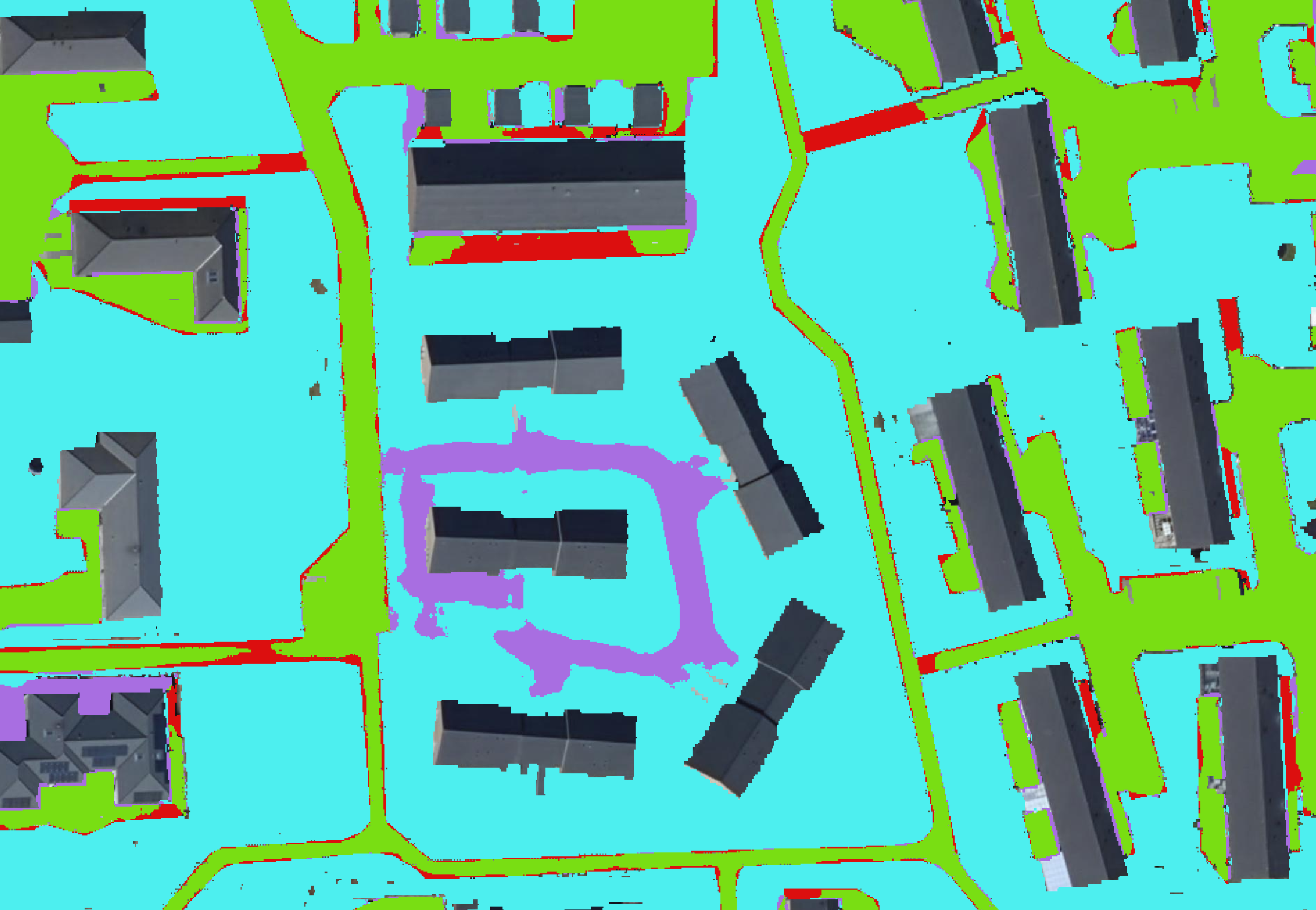}
        \subcaption{Example of label error, due to mismatch between the origin time of input features and target labels.}
    \end{minipage}
    \quad
    \begin{minipage}[t]{0.3\textwidth}
        \centering
        \includegraphics[width=\textwidth]{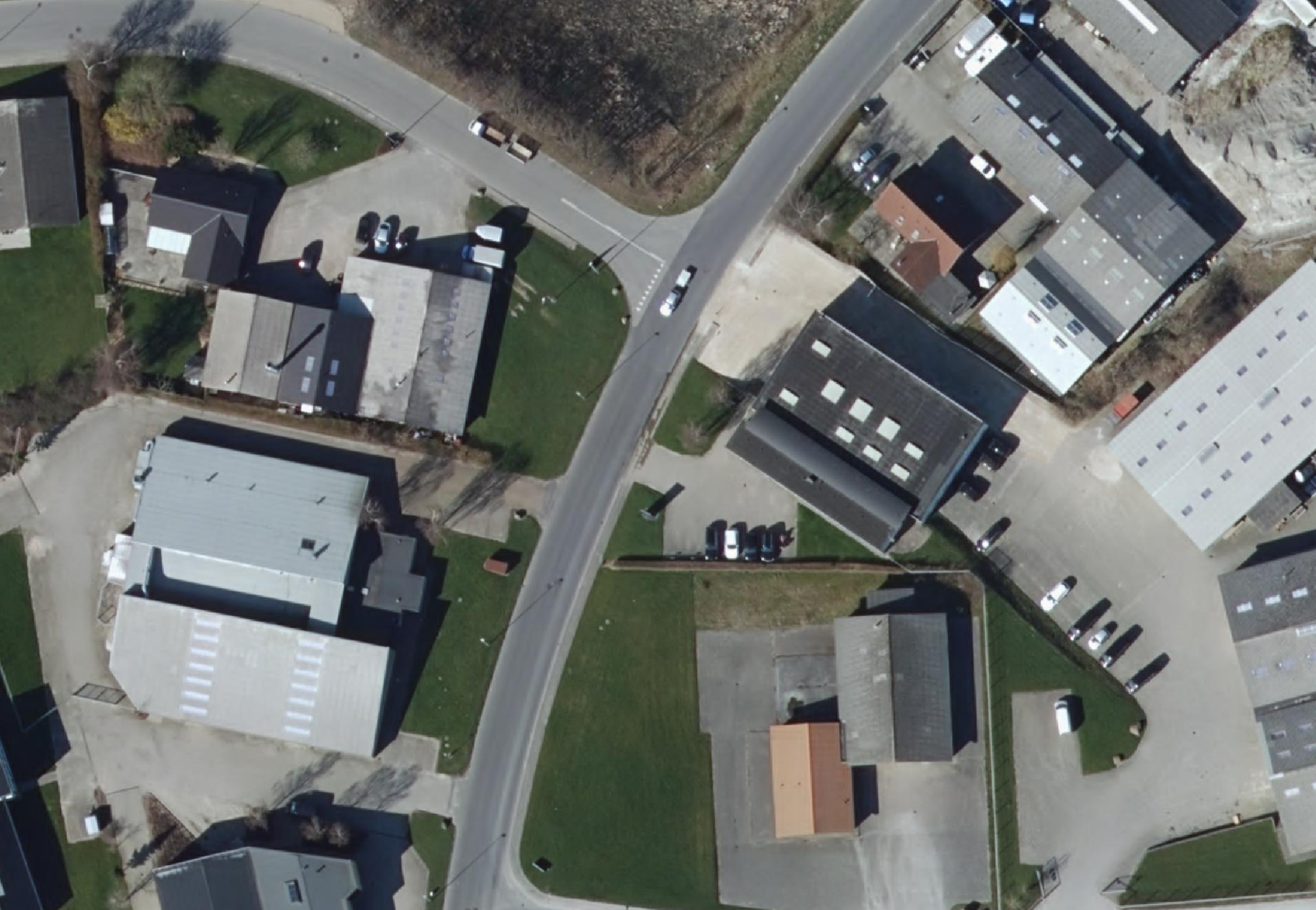}
        \includegraphics[width=\textwidth]{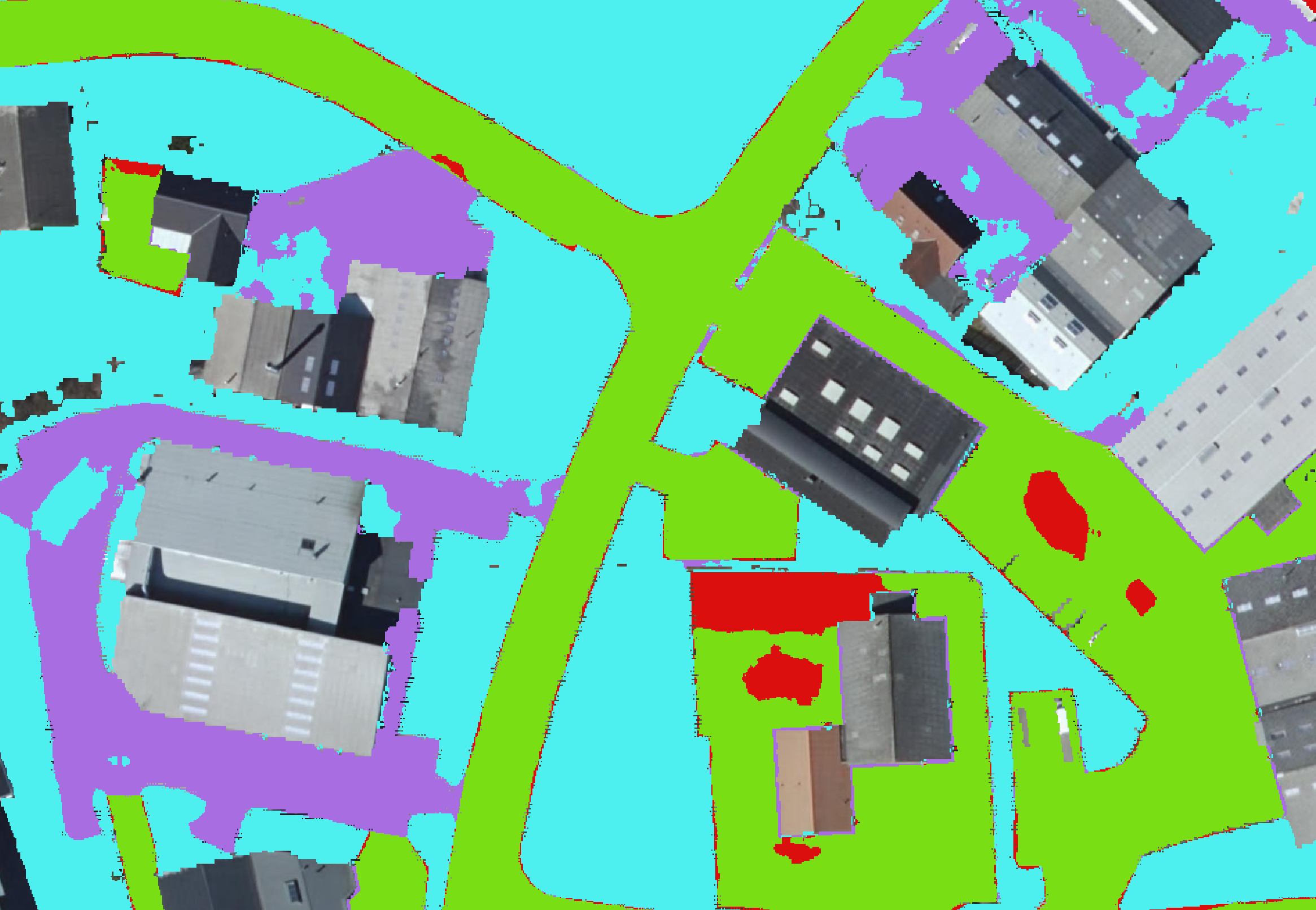}
        \subcaption{Example of 'difficult to classify' dirt road.}
    \end{minipage}
    \caption{Examples of possible misclassifications on the validation set $D_\textrm{skovby}$ using the \emph{U-Net} model.
      The coloring is as follows. True Positives are \crule{79de13}{0.2cm}, False Positives are \crule{a86ee1}{0.2cm},
      True Negatives are \crule{4defef}{0.2cm} while False Negatives are \crule{dc0f0f}{0.2cm}.}
    \label{fig:dirt}
\end{figure}
The experiments clearly show that the U-Net model has significantly better performance than the GBT model.
However,  there is still some error left which we analyze in  Table \ref{table:sep-permf} and Figure \ref{fig:dirt}.
While the performance across different types of terrain vary slightly between the models, all models have particular difficulty with unpaved road.
From Figure \ref{fig:dirt} we see  that this is not surprising, given the ambiguities present  in the data. 
Figure \ref{fig:dirt} also provides examples of possible errors in the target labels that affect performance numbers.
\begin{table}
\caption{Performance comparison for the best U-Net models considered in Table \ref{table:large_test} separated on the different classes from $P_\textrm{labels}$.
  Error part is the percentage ot total error for the classifier.
  The first row (Total Area) shows the percentage of each type of terrain in the validation set considered.
}
\begin{tabular}{lllllllll}
\toprule
Layers & Data Size & Statistic & Road & Side-Walk & Terrace & Unpaved Road & Rest\\
\midrule
& & Total Area &0.13 & 0.04 & 0.02 & 0.03 & 0.79 \\
\multirow{2}{*}{3} & \multirow{2}{*}{Small} & Accuracy & 0.91 & 0.83 & 0.84 & 0.56 & 0.99 \\
  & & Error Part & 0.25 & 0.15 & 0.06 & 0.28 & 0.23 \\
\multirow{2}{*}{3} & \multirow{2}{*}{Big} & Accuracy & 0.87 & 0.90 & 0.86 & 0.20 & 0.98 \\
&  & Error Part & 0.28 & 0.06 & 0.04 & 0.38 & 0.23 \\
\multirow{2}{*}{5} & \multirow{2}{*}{Big} & Accuracy & 0.93 & 0.90 & 0.90 & 0.50 & 0.99 \\
& & Error Part & 0.23 & 0.10 & 0.04 & 0.37 & 0.24 \\
\bottomrule
\end{tabular}
\label{table:sep-permf}
\end{table}
\begin{figure}
    \centering
    \begin{subfigure}[t]{0.45\textwidth}
      \centering
      \includegraphics[width=\textwidth]{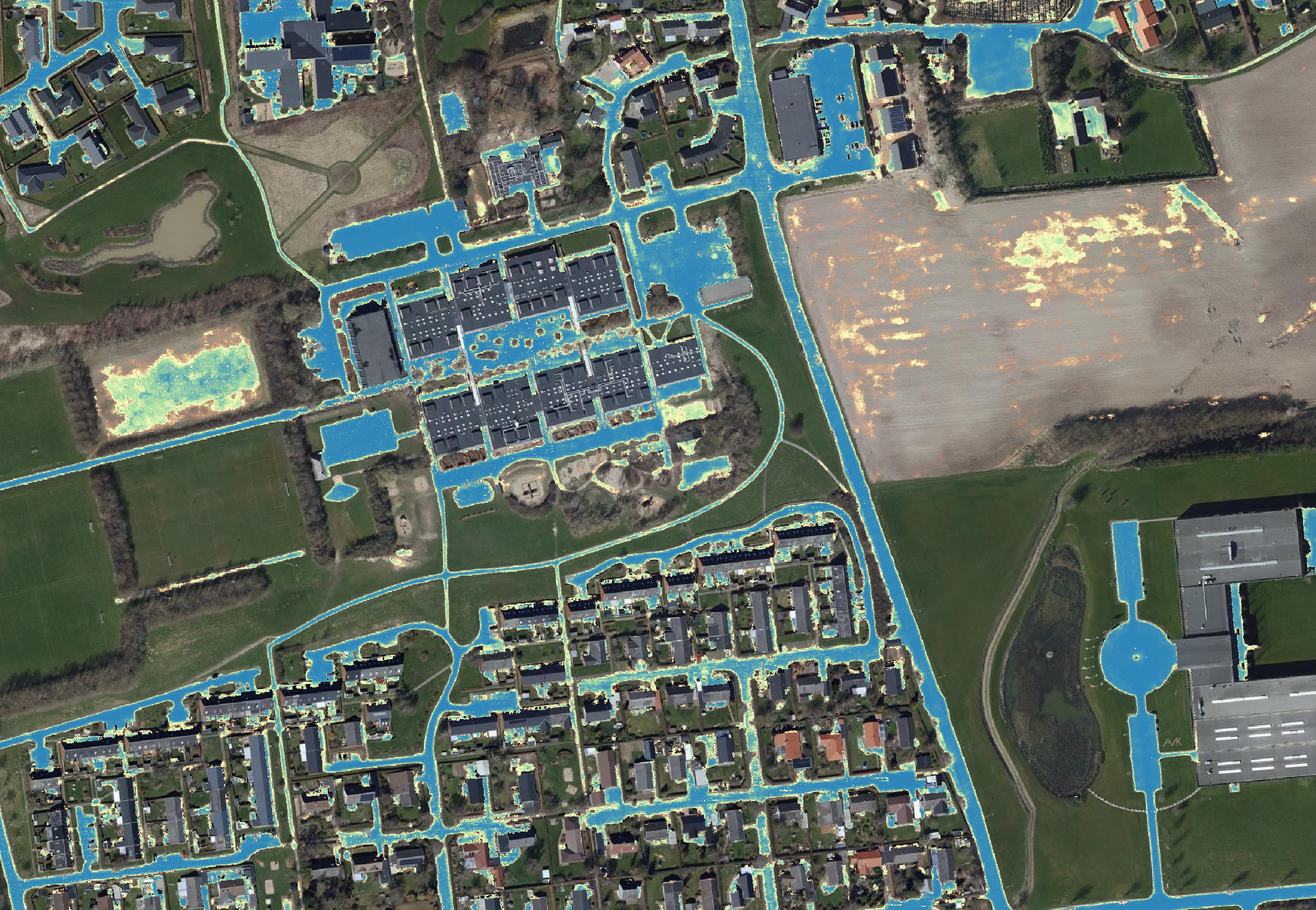}
      \caption{Heatmap of predictions using trained Gradient Boosted Trees model with encoded intensity.}
      \label{fig:gbm_no_intensity}
    \end{subfigure}
    \begin{subfigure}[t]{0.45\textwidth}
      \centering
      \includegraphics[width=\textwidth]{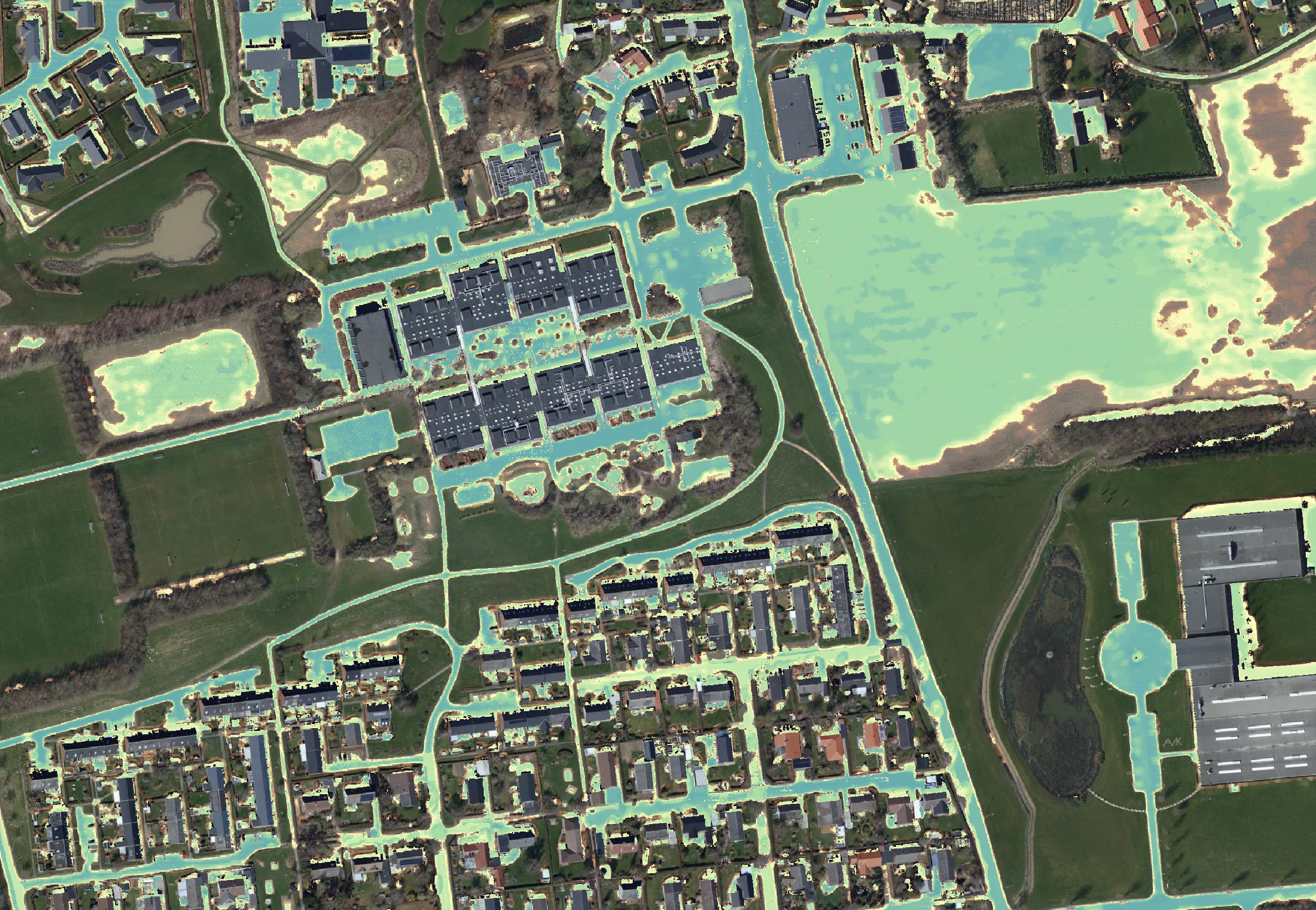}
      \caption{Heatmap of predictions using trained Gradient Boosted Trees model without encoded intensity.}
      \label{fig:gbm_with_intensity}      
    \end{subfigure}%
    \quad
    \begin{subfigure}[t]{0.45\textwidth}
      \centering
      \includegraphics[width=\textwidth]{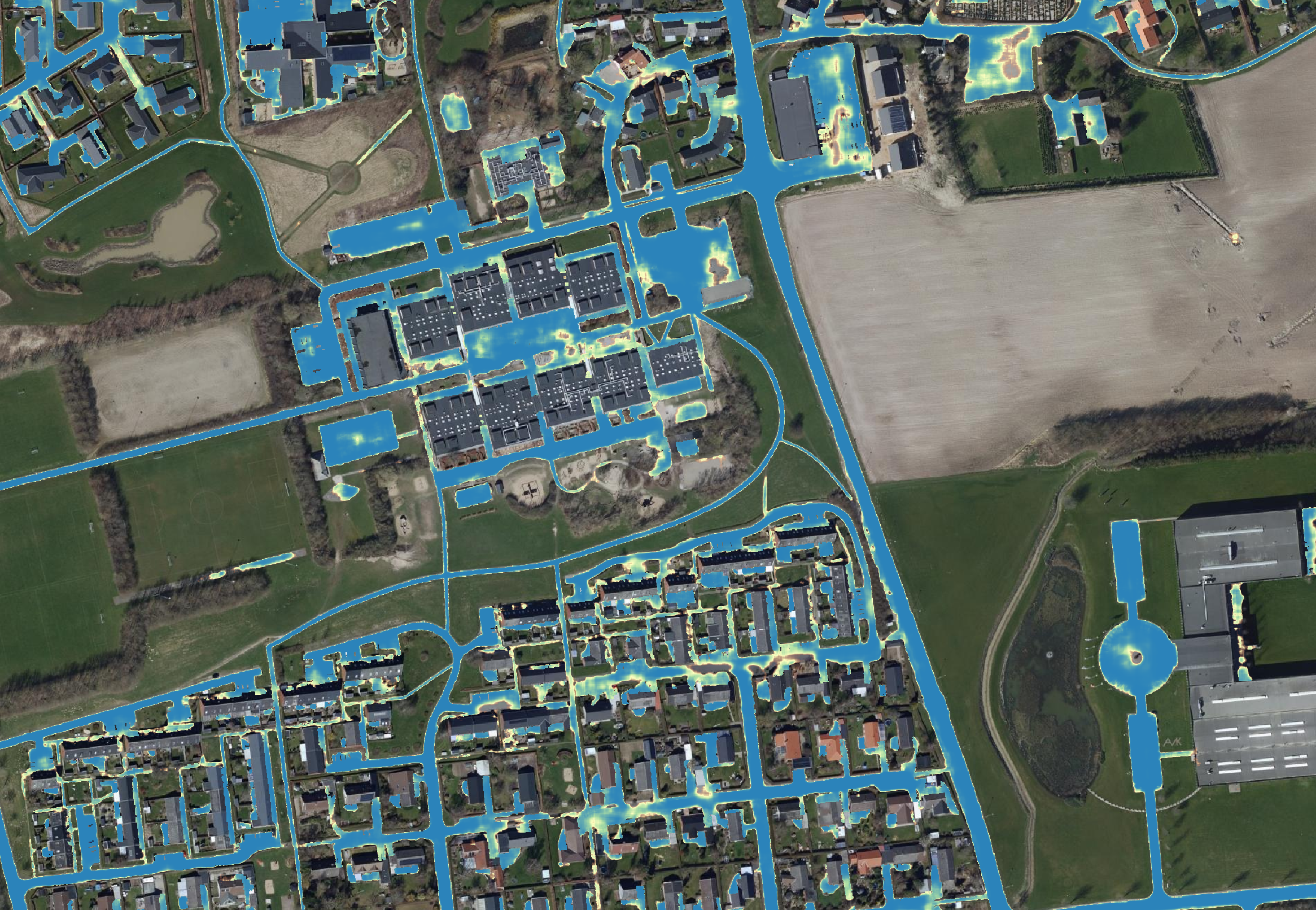}
      \caption{Heatmap of predictions using a trained U-Net model (small data, 3 layers) with encoded intensity.}
      \label{fig:unet_with_intensity}      
    \end{subfigure}
    \begin{subfigure}[t]{0.45\textwidth}
      \centering
      \includegraphics[width=\textwidth]{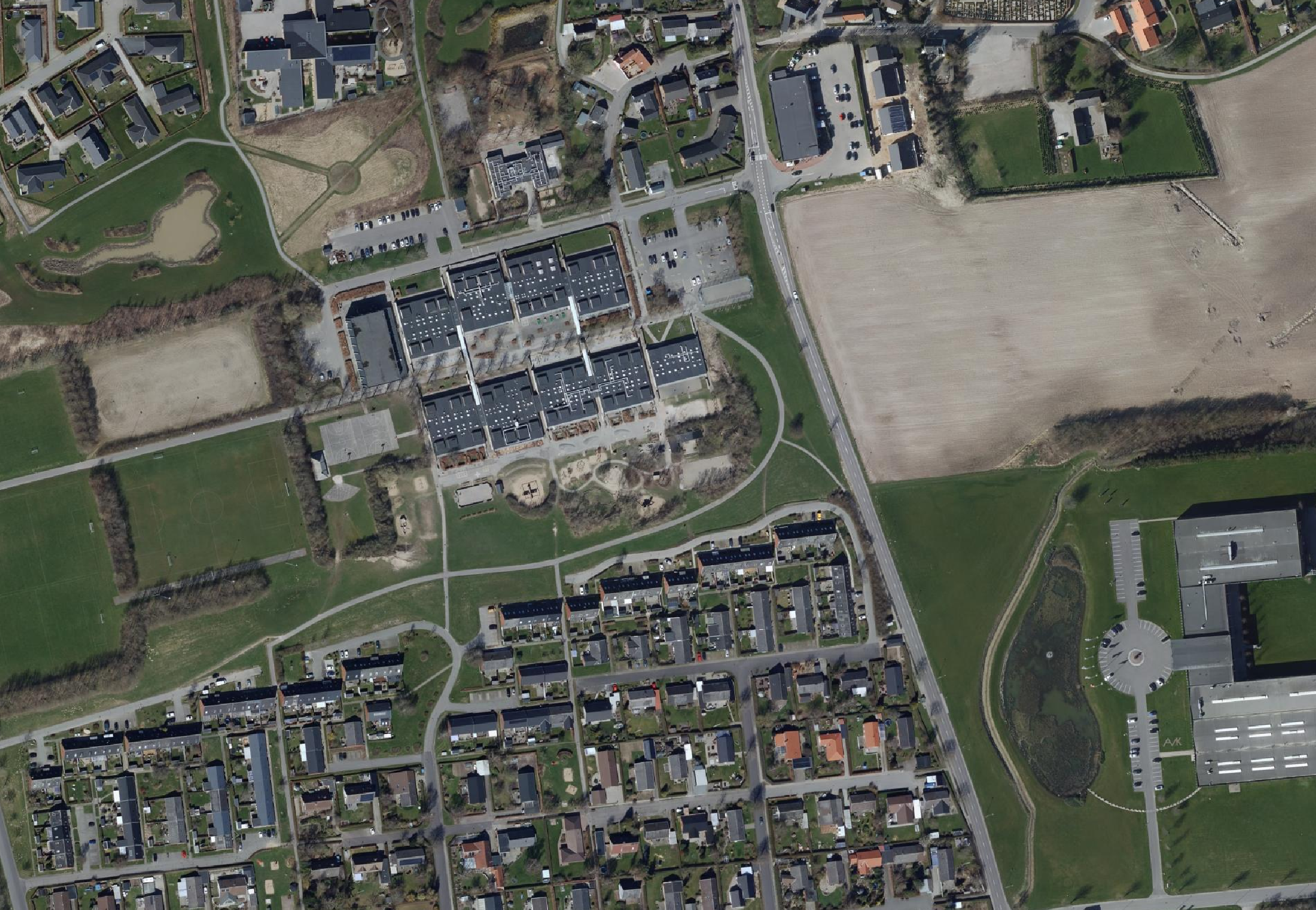}
      \caption{Orthophoto for comparison.}
      \label{fig:ortho_ref}      
    \end{subfigure}
    \caption{Overlaid heatmap of predictions from different trained models, on a crop of the validation area from $D_\textrm{skovby}$.
      The heatmap is transparent below $0.5$ and then transitions through the 3 colors \crule{ffffbf}{0.2cm}, \crule{abdda4}{0.2cm} and \crule{2b83ba}{0.2cm}.}
    \label{fig:heatmap}
\end{figure}
Finally, in Figure \ref{fig:heatmap} we show a visual comparison between the algorithms.
The figure shows that encoding the \emph{intensity} feature improves performance significantly for the GBM models, and confirms that the
U-Net based model is the superior method for this problem and that it giver very good and very reasonable predictions. 

\subsection{Things that did not help}

\subsubsection{Small Tweaks and optimizations}
We tried many variations of the intensity transformation, and while the transformed intensity feature changed with the encoding algorithm the end result did not. This may be explained by the fact that the intensity feature only has a small effect on the final result.
We also tested different numbers of channels used in the U-Net model, using different recent activation functions such as SiLU and GELU \cite{gelu}. 
We tested different variations of the hyperparameters for the Focal Loss function varying the shape of the loss function used.
However, in our experiments none of these variations had any noticeable effect on the final outcome of the algorithm.

\subsubsection{U-Net modifications}

We tried several modifications to the model architecture described in Section \ref{sec:methods}. Specifically we tried \emph{Depth-wise Seperable Convs} \cite{chollet2017xception},  \emph{Dillated convolutions} \cite{holschneider1990real} and various forms of \emph{skip-connections} \cite{he2016deep,zhou2018unet++}. 
\paragraph{Depth-wise Seperable Convolutions} This is a generalization of the standart method of applying convolutions, where each output channel depends on all input channels through a \emph{single} convolution operation.
Instead, one composes an operation in a two step process. First, for each input channel, apply a convolution kernel that only returns one output channel, and then apply a second convolution to the stacked output of these operations. This convolution uses a $1 \times 1 \times c_\textrm{out}$
kernel, where $c_\textrm{out}$ is the number of output channels, mapping the features for each cell into a different number of dimensions and producing the final output. This operation has been shown to enable significant reductions in parameter count while maintaining performance.
We experience the same effect. A reduction in the total number of parameters by approximately a factor of three, while maintaining performance. However, no increase in performance was observed.

\paragraph{Dillated convolutions and skip-connections}
This technique increases the receptive field of a convolution operation, by skipping every $i$'th pixel, while maintaining the same number of parameters. This creates a tradeoff between resolution and receptive field size, and by stacking multiple \emph{Dilated Convolutions} with different skip factors, you can create a model that has a large receptive field for every output, while keeping parameter count under control. We tried this, and achieved no increase in performance. We also tried various skip-connections, which is supposed to improve training by easing the flow of gradients.
Specifically we tried the \emph{UNet++} \cite{zhou2018unet++} architecture and variants thereof, but none of those increased performance.

\subsubsection{Graph Networks}
As described we cannot apply the U-Net directly to the \lidar point cloud data. Instead we rasterized the raw point cloud data into image form potentially losing information contained in the point cloud.
To test whether using the raw point cloud directly could improve our results, we tested the \emph{PointNet++}\cite{pointnet++} architecture, which is specifically designed for data in point cloud form.
Though we were able to achieve similar performance to our U-Net architecture, this came at a cost of larger parameter count and slower training convergence.
We suspect this is because the problem we consider is essentially 2-dimensional, and the last dimension (the height above ground),
can accurately be represented as an input channel in the data passed to the convolutional neural network, so that the added complexity of modelling the terrain as a point cloud becomes a drag instead of a gain.

In general we suspect that the failure to increase performance through architectural changes is caused by the limited quantity and quality of our training labels.
The quality or the labels used is primarily impeded by the fact that they were created over a 2 year period, and suffer from the underlying changes made in that period, creating inconsistencies between different input features and output labels.

\subsection{Things we did not try}
Another popular model for pixel segmentation is \emph{Mask R-CNN}\cite{maskrcnn}. In our problem formulation, we consider this to be inappropriate,
since the problem that Mask R-CNN tries to solve is significantly harder than our problem in the following sense.
When choosing whether a U-Net type model or a Mask R-CNN type model is appropriate, the central question is whether the instances that need to be detected overlap.
If that is the case, there needs to be some process that separates individual instances.
In Mask R-CNN this is handled through a \emph{bounding box proposal network}, which for every image pixel,
every class and a fixed number of image crop ratios, predict a probability that an instance of that class is within a particular bounding box.
Only if this probability is high enough, a feature map cropped from the original input using the bounding-box is input to a class-specific neural network that estimates the segmentation mask.
In our case, we only need that last part (the segmentation mask neural network), as our problem is binary (hence we do not need to select the right network) and overlapping instances is not an issue.

\section{Conclusions}
In this work we have shown how to construct an accurate and efficient fortification segmentation algorithm based on the U-Net neural net architecture that can be applied to large areas.
We have shown that combining data from orhophotos including near-infrared and \lidar data, including the (transformed) intensity of return feature, provides the best performance.
However using only one of the data sources still allows for high quality segmentations of fortified areas.
We have shown how to remedy the issue of different devices yielding different feature distributions in an automated and generalizable way.
Our experiments also reveal that the troublesome Intensity of Return feature may be discarded with only negligible reduction in quality. This allows the training method to be oblivious to the actual source of the device that created the each of the \lidar data points considered, which may be a considerable logistical burden. Furthermore, if this problem is fixed by the measurement system in the future incorporating the feature again may provide at least a small increase in performance.
Finally, in our experiments there was little to be gained by different architectural tweaks of the U-Net algorithm. While there may still be room for improvement by optimizing architecture, for instance applying Transformers and extensive pretraining or adding post processing of the U-Net ourput, our results suggest that generating labels of higher quality  may be the best way forward if performance needs to be improved. This may be potentially be achieved by algorithmically detecting a large portion of the bad labels. We leave such investigations for future work,

\section{Acknowledgement}
The work on this paper was proposed and started under the guidance of Professor Lars Arge, Aarhus University. Lars Arge had been fighthing cancer for several years, and had he not sadly passed away before the work was finished and the paper written, he would have been a co-author.

\bibliographystyle{abbrv}
\bibliography{main}
\end{document}